\documentclass[10pt,journal,compsoc]{IEEEtran}
\ifCLASSOPTIONcompsoc
\usepackage[nocompress]{cite}
\else
\usepackage{cite}
\fi
\ifCLASSINFOpdf
\else
\fi

\usepackage{color}
\usepackage[table]{xcolor}
\usepackage{times}
\usepackage{epsfig}
\usepackage{graphicx}
\usepackage{amsmath}
\usepackage{amssymb}
\usepackage{url}
\usepackage{booktabs}
\usepackage{makecell}
\usepackage{multirow}
\usepackage{caption}
\usepackage{gensymb}
\usepackage[export]{adjustbox}
\usepackage{threeparttable}
\usepackage[linesnumbered,lined,boxed,commentsnumbered,ruled]{algorithm2e}
\usepackage[pagebackref,breaklinks=true,colorlinks,citecolor=blue,urlcolor=blue,linkcolor=blue,bookmarks=false]{hyperref}
\usepackage{algorithmic}
\usepackage{rotating}
\usepackage[caption=false]{subfig}
\renewcommand{\paragraph}[1]{\noindent\textbf{#1}~~}

\usepackage{subfloat}
\usepackage{tabularx}
\begin{document}
		\title{Panoptic-PartFormer++: A Unified and Decoupled View for Panoptic Part Segmentation}
		
		\author{Xiangtai Li,
			Shilin Xu,
			Yibo Yang\textsuperscript{$\dagger$},
			Haobo Yuan,
			Guangliang Cheng,
			Yunhai Tong\textsuperscript{$\dagger$},\\
			Zhouchen Lin, 
			Ming-Hsuan Yang, 
			Dacheng Tao
			
			\IEEEcompsocitemizethanks{\IEEEcompsocthanksitem X.~Li, S.~Xu, Y.~Tong, and Z.~Lin are with National Key Laboratory of General Artificial Intelligence, Peking Univeristy, Beijing, China. This work is supported by the National Key Research and Development Program of China (No.2023YFC3807600). The first two authors contribute equally to this extension.  
				\IEEEcompsocthanksitem Y.~Yang is with KAUST, Jeddah, KSA.
				\IEEEcompsocthanksitem G.~Cheng is with the Department of Computer Science at the University of Liverpool, UK.
				\IEEEcompsocthanksitem M.-H. Yang and H.~Yuan are with the Department of Computer Science and Engineering at the University of California, Merced, US.
				\IEEEcompsocthanksitem D.~Tao is with Nanyang Technological University, Singapore.
				\IEEEcompsocthanksitem Corresponding to: Y.~Tong and Y.~Yang
			}% <-
			%\thanks{The first two authors contribute equally.}
		}

	\IEEEtitleabstractindextext{
\begin{abstract}
Panoptic Part Segmentation (PPS) unifies panoptic and part segmentation into one task. Previous works utilize separate approaches to handle things, stuff, and part predictions without shared computation and task association. We aim to unify these tasks at the architectural level, designing the first end-to-end unified framework, Panoptic-PartFormer. Moreover, we find the previous metric PartPQ biases to PQ. 
To handle both issues, we first design a meta-architecture that decouples part features and things/stuff features, respectively. We model things, stuff, and parts as object queries and directly learn to optimize all three forms of prediction as a unified mask prediction and classification problem. We term our model as Panoptic-PartFormer. 
Second, we propose a new metric Part-Whole Quality (PWQ), better to measure this task from pixel-region and part-whole perspectives. It also decouples the errors for part segmentation and panoptic segmentation. 
Third, inspired by Mask2Former, based on our meta-architecture, we propose Panoptic-PartFormer++ and design a new part-whole cross-attention scheme to boost part segmentation qualities further. We design a new part-whole interaction method using masked cross attention. 
Finally, extensive ablation studies and analysis demonstrate the effectiveness of both Panoptic-PartFormer and Panoptic-PartFormer++. Compared with previous Panoptic-PartFormer, our Panoptic-PartFormer++ achieves 2\% PartPQ and 3\% PWQ improvements on the Cityscapes PPS dataset and 5\% PartPQ on the Pascal Context PPS dataset. On both datasets, Panoptic-PartFormer++ achieves new state-of-the-art results. Our models can serve as a strong baseline and aid future research in PPS. 
The source code and trained models will be available at~\url{https://github.com/lxtGH/Panoptic-PartFormer}.
\end{abstract}

}

	\maketitle

	% is not selected <OR> if conference mode is selected - because compsoc
	% conference papers position the abstract like regular (non-compsoc)
	\IEEEdisplaynontitleabstractindextext

	% \fi
	\IEEEpeerreviewmaketitle
	
	% text~\cite{woo2018cbam}
	\section{Introduction}
% logic:
% 1, importance of understanding both scene and part segmentation, why we needs such task.
% 2, previous work overview. Could we build a unified and end-to-end framework   
% 3, The motivation and insights of panoptic-partformer 
% 4, methods details description of panoptic-partformer 
% 5, limitation of panoptic partformer and limitation of this tasks. 
% 6, our new contribution. 
% 7, experiment results.

\IEEEPARstart{O}{ne} fundamental problem in computer vision is to understand a scene at \textit{multiple levels of concepts}. In particular, when people look at a figure, they can not only catch each visual entity, such as cars, buses, and background contexts like sky or road, but also understand the parts of entities, such as person-head and car-wheel, etc. The former is named as \textit{scene parsing}, while the latter is termed as \textit{part parsing}. One representative direction of unified scene parsing is Panoptic Segmentation (PS)~\cite{kirillov2019panoptic}. It predicts a class label and an instance ID for each pixel and unifies the foreground objects (named as things) and background context (named as stuff). Part parsing has many cases, such as human or animal parts or car parts~\cite{liang2015human,geng2021part}. 
%
%Both directions are independent, while both are equally important for many vision systems, including autonomous driving and robot navigation~\cite{cordts2016cityscapes}.
While these tasks are often addressed in separate contents, both are equally important for numerous vision systems, including autonomous driving and robot navigation~\cite{cordts2016cityscapes}.

\begin{figure}[!t]
	\centering
	\includegraphics[width=0.90\linewidth]{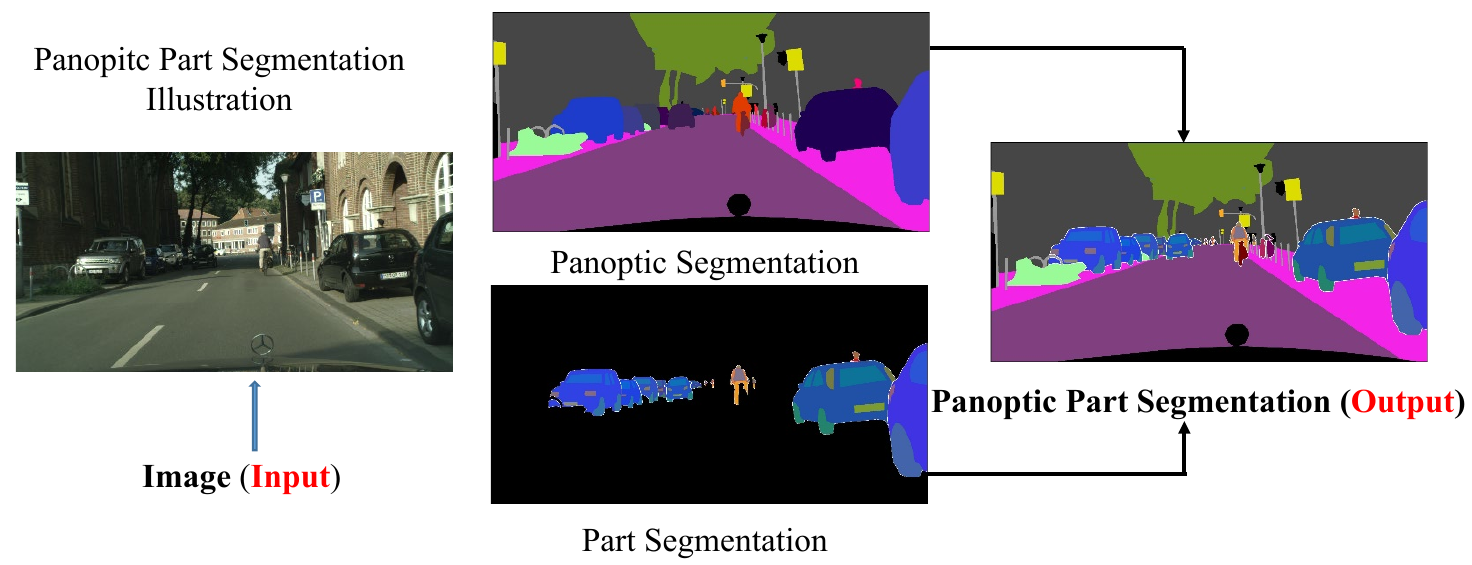}
	\caption{Illustration of the Panoptic Part Segmentation task. It combines Panoptic Segmentation and Part Segmentation in a unified manner that provides a multi-level concept understanding of the image. Best viewed in color.}
	\label{fig:teaser_01}
\end{figure}

\begin{figure}[!t]
	\centering
	\includegraphics[width=0.90\linewidth]{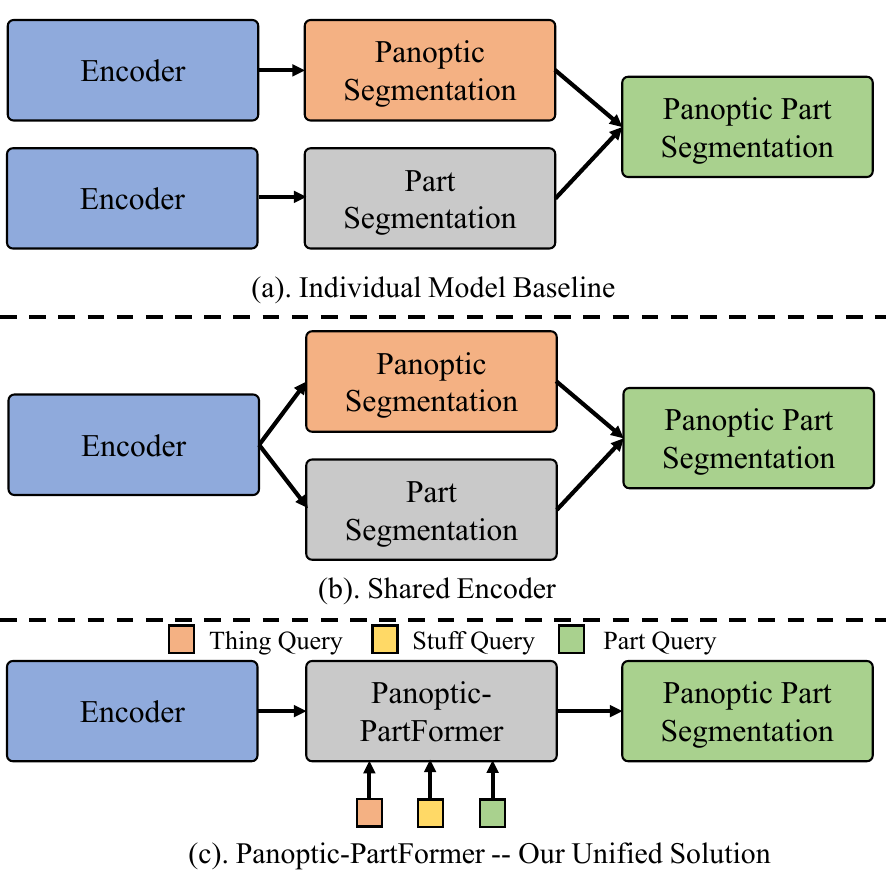}
	\caption{(a) The baseline method proposed in~\cite{degeus2021panopticparts} combines results of panoptic segmentation and part segmentation. (b) Panoptic-FPN-like baseline~\cite{kirillov2019panopticfpn,li2019attention,xiong2019upsnet} adds part segmentation into the current panoptic segmentation frameworks. (c) Our proposed approach represents things, stuff, and parts via object queries and performs joint learning in a unified manner.}
    \label{fig:teaser_02}
\end{figure}

Recently, the Panoptic Part Segmentation (PPS, or Part-aware Panoptic Segmentation)~\cite{degeus2021panopticparts} is proposed to unify these multiple levels of abstraction into one single task. As shown in Fig.~\ref{fig:teaser_01}, it requires a model to output per-pixel scene-level classification for background stuff, segment things into individual instances, and segment each instance into specific parts. 
Several approaches~\cite{maskrcnn,zhao2017pyramid,Zhao2019BSANet} with hybrid formulations have been proposed to tackle this task. 
As shown in Fig.~\ref{fig:teaser_02}(a), these methods fuse individual predictions for different tasks to obtain the PPS results,
%different individual model predictions to obtain PPS results, 
\emph{i.e.}, obtaining panoptic and part segmentation results individually based on separate models and then merging them for the PPS task.
This makes the entire process complex, with substantial engineering efforts. 
In addition, the shared computation and task association are ignored, which leads to huge costs and inferior results. Another solution to this task is to make the part segmentation an extra head with a shared backbone~\cite{jagadeesh2022multi}, as shown in Fig.~\ref{fig:teaser_02}(b). 
Such design is well explored in the early PS studies~\cite{xiong2019upsnet,kirillov2019panopticfpn,li2020panopticFCN,chen2020banet,li2018learning,porzi2019seamless,yang2019sognet,Wu2020AutoPanopticCM,cheng2020panoptic,axialDeeplab}. 
However, most of them treat PS as separated tasks~\cite{xiong2019upsnet} or sequential tasks with post-processing components~\cite{cheng2020panoptic}, which still cannot explore the mutual effect of the scene-level and part-level understandings.

Since Detection Transformer (DETR)~\cite{detr}, there have been several works~\cite{wang2020maxDeeplab,cheng2021maskformer,zhang2021knet,cheng2021mask2former} unifying both thing and stuff learning via \textit{object queries} in PS, which makes the entire pipeline elegant and achieves strong results. 
%where the mask classification and prediction can be performed directly. 
%
These results show that many complex components, including NMS and box detection heads, can be removed, and mask classification and prediction can be performed directly.
In particular, such designs consider the complete scene understanding via performing interactions among object queries and image features. 
%things, stuff, and part simultaneously. 
%
Joint training with things, stuff, and parts likely leads to better part segmentation results since the complete scene information renders more discriminative information, such as global context. 
%Moreover, adding such global context also benefits the local part segmentation. For example, the car parts are different with human parts.
%{\color{red} On the other hand, why panoptic segmentation benefits ?}
%from the more precise local context offered by part representation. 
%
Motivated by this, we propose a unified model to tackle the PPS task.

This work presents a simple yet effective framework named Panopic-PartFormer, a unified model for PPS tasks. 
As shown in Fig.~\ref{fig:teaser_02}(c), we introduce three different queries for modeling thing, stuff, and part predictions, respectively. 
Then, a decoupled decoder is proposed to generate fine-grained features that decode things, stuff, and part mask predictions. 
%
%The decoupled decoder contains a part decoder and a scene decoder. 
%
%For the part decoder, we design a feature-aligned decoder to keep more fine details in parts. 
Rather than directly using the pixel-level self-attention in Transformer, we consider the recent work K-Net~\cite{zhang2021knet} that performs self-attention on the instance level for panoptic segmentation. 
Specifically, we focus on refining queries via their corresponding query features, which are defined as the masked decoder features using the mask associated with each query. 
%. We define the \textit{query feature} as \textit{grouped features} that are generated from the \textit{corresponding mask} of each query and \textit{decoder features}. 
%
We generate masks via dot product between queries and decoder features and 
perform updating object queries with the query features. 
This operation is implemented with one dynamic convolution~\cite{zhang2021knet,peize2020sparse} and multi-head self-attention layers~\cite{vaswani2017attention} between query and query features iteratively. 
The former poses instance-wise information from features to enhance query learning.
%where the features generate the parameters. 
The latter performs inner reasoning among different queries to model the relationship among thing, stuff, and part. Moreover, the entire procedure avoids the pixel-level computation in other vision Transformer decoders~\cite{detr,cheng2021maskformer}. 
Extensive experiments (Sec.\ref{sec:ablation}) show that our approach achieves much better results than the previous designs in Fig.~\ref{fig:teaser_02} (a) and (b). It achieves the new state-of-the-art results on two challenging PPS benchmarks, including the Pascal Context PPS dataset (PPP) with about 6-7\% PartPQ gain on ResNet101, 10\% PartPQ gains using Swin Transformer~\cite{liu2021swin}, and the Cityscapes PPS dataset (CPP), with about 1-2\% PartPQ gain.

Beyond Panoptic-PartFormer framework, published in ECCV 2022~\cite{li2022panopticpartformer}, we make more significant contributions in this work. 
\textbf{First}, by analyzing the PartPQ metric in PPS, we discover that this metric unfairly overemphasizes the effect of panoptic segmentation. To address this issue, we propose a new metric called Part-Whole Quality (PWQ), inspired by the principles of part-whole modeling and pixel-wise segmentation quality. The PWQ metric evaluates segmentation quality in two ways: first, by considering both pixel-wise and region-wise segmentation quality, and second, by using a hierarchical approach that equally weighs part and scene results. The PWQ metric decouples part segmentation errors from scene segmentation errors and distinguishes between pixel-wise segmentation (mIoU) and region-wise segmentation (PQ). By introducing the PWQ metric, we hope to advance the PPS field by providing a fair evaluation of part and scene segmentation quality at both the pixel-region level and the part-whole level, while also balancing the ratio of PQ in PartPQ.

%by analyzing the PartPQ metric of PPS, we find that this metric unfairly highlights the effect of panoptic segmentation. Inspired by the part-whole modeling and pixel-wise segmentation quality, we present a new metric named Part-Whole Quality (PWQ). It contains two aspects: one from both pixel-wise and region-wise segmentation quality, the other from the hierarchical view that jointly considers part and scene results equally. The new metric decouples both part segmentation errors and scene segmentation errors. It also decouples pixel-wise segmentation (mIoU) and region-wise segmentation (PQ). With the newly proposed metric PWQ,  we hope to advance the PPS field by fairly considering part and scene segmentation from two aspects, including pixel-region level and part-whole level evaluations. It also balances the ratio of PQ in PartPQ. 

%

%
\textbf{Second}, we present a new enhanced model named Panoptic-PartFormer++. In particular, we have \textit{three} improvements over the original Panoptic-PartFormer. \textbf{1.} Rather than using the coarse masked pooling for query-level reasoning, we adopt the recent stronger baseline Mask2Former design by replacing query learning with masked cross attention of both queries and multiscale features. Since most part objects in both PPP and CPP datasets are tiny, masked pooling ignores the fine details of the part. We present a global-first strategy. In particular, we first perform the global-part masked cross attention, where the part, thing, and stuff queries are jointly learned first. Then, we adopt part masked cross attention using the part query from the previous step to learn a better part feature. 
\textbf{2.} We present an enhanced version of the decoupled decoder. We append extra semantic segmentation loss to enhance part features learning. 
Moreover, we adopt a deformable encoder to extract multiscale features
in a way similar to  Mask2Former~\cite{cheng2021mask2former}.
\textbf{3.} We jointly perform thing, stuff, and part query learning for each scale. We only supervise the part query learning for the highest resolution to keep fine-grained information. 

\textbf{Third}, we present a more detailed analysis of our newly proposed architecture and metric. In particular, we verify the effectiveness of different architectural designs in the experiment part. Our model can directly output thing, stuff, and part segmentation predictions in a box-free and NMS-free manner. A sub-task of PPS, such as Panoptic segmentation, can also be evaluated. In the experiment part, we verify our panoptic segmentation predictions on the Cityscapes dataset~\cite{cordts2016cityscapes}, and achieve better results than the recent work~\cite{li2020panopticFCN}.
Moreover, we explore the recent stronger backbones on both CPP and PPP datasets. Our final proposed Panoptic-PartFormer++ achieves the new state-of-the-art results on three metrics, including PQ, PartPQ, and our proposed PWQ. 
In particular, it achieves 63.1\% PartPQ on the CPP dataset. Compared to the previous conference version, we observe about \textbf{1\%-2\%} PartPQ and PWQ improvements. When scaling up our model, it also achieves 53.2\% PartPQ on the PPP dataset, which is more than 15\% improvement over the previous baselines and 5\% improvement over the Panoptic-PartFormer in our conference version.

	\section{Related Work}
\noindent
\textbf{Part Segmentation.} Current research on part segmentation can be divided into two categories: supervised learning and unsupervised learning. 
Supervised learning approaches for both instance and semantic part segmentation typically rely on human analysis \cite{qi2018learning}. 
In contrast, recent works on unsupervised learning~\cite{liu2021unsupervised,choudhury2021unsupervised} focus on single objects with multiple parts and propose various contrastive learning approaches. 
However, most of these works do not consider multiple objects and background context. Therefore, in this paper, we mainly review supervised learning approaches. 
Several works \cite{fang2018weakly,wang2019CNIF} have designed specific methods for semantic part segmentation, which only considers per-pixel classification. These methods focus on modeling global-part context more effectively. For human instance part segmentation, there are two paradigms: top-down pipelines \cite{li2017holistic,yang2019parsing,ji2019learning}, and bottom-up pipelines \cite{gong2018instance,zhao2018understanding,zhou2021differentiable}. 
Top-down methods use two-stage detectors \cite{maskrcnn} to jointly detect and segment each instance. For each detected instance, they perform semantic part segmentation. 
In contrast, bottom-up methods first segment each human and then perform part grouping from each detected human. Several works also focus on task-specific part segmentation, such as car part segmentation \cite{Zhao2019BSANet,michieli2020gmnet}. However, each of these settings requires a specific design. 
In comparison, this paper focuses on solving the PPS task, which naturally contains part segmentation as a sub-task.

\noindent
\textbf{Panoptic Segmentation.} Earlier work~\cite{kirillov2019panoptic} performs segmentation of things and stuff via separated networks~\cite{maskrcnn,deeplabv3plus}, and then directly combines the predictions of things and stuff for the panoptic segmentation result.
To alleviate the computation cost, recent works~\cite{kirillov2019panopticfpn,yang2019sognet} are proposed to model both stuff segmentation and thing segmentation in one model with shared backbone and different task heads. Detection-based methods~\cite{xiong2019upsnet,kirillov2019panopticfpn} usually represent things with the box prediction, while several bottom-up models~\cite{cheng2020panoptic,axialDeeplab} perform grouping instance via pixel-level affinity or center heat maps from semantic segmentation results. 
Several recent works~\cite{qiao2021detectors,hou2020real} focus on designing a stronger backbone and more effective task association method to merge semantic and instance segmentation. However, the former introduces a complex process, while the latter suffers from a performance drop in complex scenarios. Recently, several works~\cite{wang2020maxDeeplab,li2022panopticsegformer,yuan2022polyphonicformer,yu2022kmaxdeeplab} propose to obtain panoptic segmentation masks without box supervision directly. However, these works do \textbf{\textit{not}} cover the \textit{knowledge of part-level semantics} of images, which can provide more comprehensive and hierarchical information for scene understanding.

\noindent
\textbf{Part-Whole Modeling.} Part-Whole modeling has a long history in computer vision. Before the deep learning era, part-based methods~\cite{felzenszwalb2010object,fergus2003object} are mainly used for object detection and recognition. Several approaches use part discovery in the deep learning era as essential cues for the fine-grained classification~\cite{zhang2013deformable}. There are also several works~\cite{tritrong2021repurposing,zhang2021datasetgan} using generative adversarial networks for few-shot part segmentation. Meanwhile, several works~\cite{sabour2021unsupervised,xu2019unsupervised} explore the part motion information in dynamic video inputs.
These works mainly focus on the parts of a single object for recognition, which ignores the background context or multiple complex objects. 
To better understand the entire scene and unify the instance-wise part segmentation~\cite{yang2020eccv,gong2018instance}, the PPS task~\cite{degeus2021panopticparts} is proposed. This work annotates two datasets (Cityscape PPS~\cite{cordts2016cityscapes} and Pascal Context PPS~\cite{Everingham2010Pascal}) and proposes a new metric named PartPQ~\cite{degeus2021panopticparts,kirillov2019panoptic} for evaluation. This work also presents several baseline methods to obtain the final results. 
However, all these methods use separated networks to obtain the panoptic segmentation result.
Besides, they use existing panoptic segmentation algorithms with part semantic segmentation as an isolated subnetwork. Recently, another work~\cite{tang2022visual} formulated the PPS task as multi-level recognition by request. However, it still uses the two separated models to handle part and thing segmentation. As a comparison, we focus on the much more complex scene understanding task, considering multiple instances with their parts. We aim to design a \textit{simple and unified} network for all the sub-tasks.

\noindent
\textbf{Vision Transformer (ViT).} 
ViTs have recently been explored in numerous vision problems, including representation learning as a feature extractor, vision-language modeling, and using object queries for downstream tasks. %detection-related tasks. 
For the first aspect, ViTs~\cite{VIT,liu2021swin,deit_vit,zhang2022eatformer} have more advantages in modeling global-range relations among the image patch features. Most recent works~\cite{li2022uniformer,guo2021cmt} combine the local CNN design with ViTs. 
Moreover, recent work~\cite{MaskedAutoencoders2021} also adopts ViTs for self-supervised learning via masked images, achieving stronger results than supervised learning. 
For the second aspect, several works~\cite{CLIP} explore large-scale text-image pairs to build Vision Language Models (VLMs), which can be used for many zero-shot or open vocabulary settings~\cite{OpenSeg}. 
Moreover, there are several works~\cite{chen2022vitadapter} transferring or adapting the knowledge of Transformer for downstream tasks. 
The third application uses the object query proposed by DETR~\cite{detr} to unify or simplify the task pipeline. It models the object detection task as a set prediction problem with learnable queries. 
Other works~\cite{peize2020sparse,zhu2020deformabledetr} explore the locality of the learning process or location prior to refine the object query. 
Query-based learning can also be applied to more fields, including instance segmentation~\cite{dong2021solq,xu2022fashionformer}, video object detection~\cite{zhou2022transvod}, object tracking~\cite{meinhardt2021trackformer}, and video segmentation~\cite{li2022videoknet}. 
Our method is inspired by these works, enjoying the benefit of ViTs backbone to unify and simplify the PPS task based on query learning.
To our knowledge, we propose the \textit{first unified Transformer model} for the PPS task.

\noindent
\textbf{Multi-Task Dense Prediction.} This task~\cite{densePredic_dac_2018,Pad-net_2018,PAP-Net_2019} aims to predict pixel-level outputs, including semantic segmentation, depth estimation, surface normal, and object boundary of the scene, simultaneously. Earlier works~\cite{densePredic_cwkdis_2021,densePredic_dcmd_2021,pattern_struct_diffusion_2020,multitask_mtst_2021,multitask_UM_2019,Mti-net_2020} explore cross-task learning by co-propagating intra-task and inter-task structures with different fusion strategies, such as dense fusion and cross-task affinity modeling. 
Recently, several concurrent works~\cite{invpt2022,xu2023mqformer,taskprompter2023} have also explored transformer architecture to leverage the task association. 
In particular, MQtransformer~\cite{xu2023mqformer} adopts task-specific queries to learn the correlation of different task features, while TaskPrompter~\cite{taskprompter2023} designs a spatial-channel task prompt to interact with patch tokens along spatial and channel dimensions. 
These methods output pixel-level format of the input image without considering instance discrimination and part-whole correlation between object and parts. 
Thus, via our extensive experiment in Sec.~\ref{sec:ablation_pppformer_plus}, their approaches cannot obtain performance gains on PPS. This is because these methods 
only explore feature-level interaction and ignore entity-level interaction. 
However, we argue that entity-level interaction is more important, as proved in both PanopticPartFormer and PanopticPartFormer++ (Sec.~\ref{sec:ablation_pppformer} and Sec.~\ref{sec:ablation_pppformer_plus}).

	\section{Method}
% logic:
% 1. Problem Reformulation and related works.
% 1-2 -> Meta architecture 
% 2. Previous PanopticPartFormer 

% 3. Existing problems of PPS using PPFormer/ Drawbacks of PanopticPartFormer 

% 4. A new proposed metric of PPS task.

% 5. New proposed PanopticPartFormer++ for PPS.

% 6. Training and Inference Details. (Difference of PPFormer and PPFormer++)

% We first review the definition of PPS and the recent Mask Transformers for segmentation as preliminary in Sec.~\ref{sec:background}. Then, we discuss the potential issues of the PPS metric and propose a new balanced metric named Part-Whole Quality in Sec.~\ref{sec:PWQ}. Next, we describe the meta-architecture of PanopticPartFormer in Sec.~\ref{sec:meta} and detail the previous PanopticPartFormer design~\cite{li2022panopticpartformer} in Sec.~\ref{sec:ppformer_eccv} as a unified model. 
% %
% In Sec.~\ref{sec:ppformer++}, we present a new enhanced version named PanopticPartFormer++. 
% %
% Finally, we describe the training and inference procedure of both models in Sec.~\ref{sec:train_and_inference}. 

We first review the definition of PPS and the recent Mask Transformers for segmentation as preliminary in Sec.~\ref{sec:background}. Then, we describe the meta-architecture of PanopticPartFormer in Sec.~\ref{sec:meta} and detail the previous PanopticPartFormer design~\cite{li2022panopticpartformer} in Sec.~\ref{sec:ppformer_eccv} as a unified model. 
Via our PanopticPartFormer, we discuss the potential issues of the PPS metric and propose a new balanced metric named Part-Whole Quality in Sec.~\ref{sec:PWQ}. Next, in Sec.~\ref{sec:ppformer++}, we present a new enhanced version named PanopticPartFormer++ further to enhance our proposed PWQ metric and previous PartPQ metric. Finally, we describe the training and inference procedure of both models in Sec.~\ref{sec:train_and_inference}.

\subsection{Preliminaries}
\label{sec:background}

\noindent
\textbf{Problem Definition.} Given an input image $ I \in \mathbb{R}^{H\times {W}\times 3}$, where ${H}$ and ${W}$ are the height and width of the image, the goal of PPS is to output a group of non-overlapping masks $\{y_i\}_{i=1}^G = \{(m_i, c_i)\}_{i=1}^G \,$ where $c_i$ denotes the ground truth class label of the mask $m_i$ and $G$ is the number of masks which depends on the input scene. In particular, $c_i \in \mathcal{L} = \{\mathcal{C}_{st}, \mathcal{C}_{th}, \mathcal{C}_{pt}\}$ may be the label of \textit{part}, \textit{thing} or \textit{stuff} and the part masks are within the thing masks. $\mathcal{C}_{st}$, $\mathcal{C}_{th}$, and $\mathcal{C}_{pt}$ are the categories of thing, stuff, and part. $\mathcal{L}$ is the whole label set. The above is defined from the region level. At the same time, each pixel is assigned a unique triple tuple from the pixel level, including semantic ID, instance ID, and part ID. If we remove the part class of PPS, this task reduces to panoptic segmentation. 
Thus, panoptic segmentation can be deemed as a specific case of PPS. We term thing, stuff segments as \textit{scene segments}.
Previous works~\cite{degeus2021panopticparts} merge individual outputs of panoptic segmentation and part segmentation, which ignores the natural part-whole relationship.

\noindent
\textbf{PartPQ Metric.} PartPQ is proposed to unify both panoptic segmentation and part segmentation~\cite{degeus2021panopticparts} and is defined as: 
\begin{equation}
    \textrm{PartPQ} = \frac{\sum_{(p,g) \in \textit{TP}}\textrm{IoU}(p,g)}{|\textit{TP}| + \frac{1}{2}|\textit{FP}|+ \frac{1}{2}|\textit{FN}|},
 \label{equ:partPQ}
\end{equation}
where ${TP}$ is a true positive segment, ${FP}$ is a false positive segment, and ${FN}$ is a false negative segment. 
This metric is based on the Intersection over Union (IoU) between a predicted segment $p$ and a ground-truth segment $g$ for one class $l$ (where $l \in \mathcal{L}$). 
A prediction is a ${TP}$ if it has an overlap with a ground-truth segment and the $\textrm{IoU} > 0.5$. An ${FP}$ is a predicted segment not matched with the ground truth, and an ${FN}$ is a ground-truth segment not matched with a prediction. $\textrm{IoU}$ contains two cases (part and non-part): 
$$ \textrm{IoU}(p,g) =
    \begin{cases}
      \textrm{mean IoU\textsubscript{part}}(p,g), & \textrm{$l \in \mathcal{L}^\text{parts}$}\\
      \textrm{IoU\textsubscript{scene}}(p,g), & \textrm{$l \in \mathcal{L}^\text{no-parts}$}
    \end{cases}$$
If a segment contains the part annotations, PartPQ calculates the mean IoU of each part. If all the things have no part annotations, the PartPQ degrades to PQ. 

\noindent
\textbf{General Mask Transformers.} Recent methods for unified segmentation adopt formulations similar to DETR~\cite{detr}. 
In Max-Deeplab~\cite{wang2020maxDeeplab}, the object queries (including thing queries and stuff queries) are used to \textit{perform cross-reasoning with high-resolution features and self-attention among the reasoned queries}.
Different methods usually have different designs for query reasoning. 
In particular, K-Net~\cite{zhang2021knet} proposes to use mask pooling with a dynamic convolution. 
Mask2Former~\cite{cheng2021mask2former} proposes masked cross attention with multiscale features. 
After this step, the updated queries are used for mask classification via an MLP. 
Like DETR~\cite{detr}, the corresponding masks are obtained by multiplying the updated queries with features. These steps are performed multiple times in a cascaded manner. 
In addition, the intermediate outputs are also supervised with mask labels during training.

\subsection{Meta Architecture For PPS}
\label{sec:meta}
\begin{figure}[!t]
	\centering
        \includegraphics[width=0.75\linewidth]{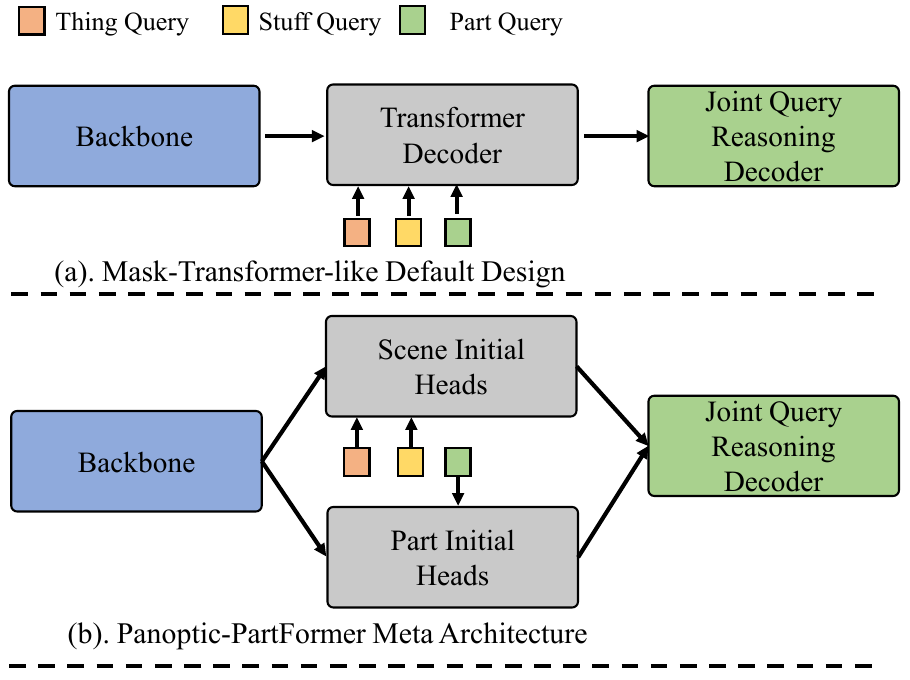}
	\caption{Meta architecture of Panoptic-PartFormer. (a) is the default general Mask Transformer design for PPS. (b) is our Panoptic-PartFormer that decouples the Scene and Part heads in cases for both learning part features and cross-reasoning stages.}
    \label{fig:meta_architecture}
\end{figure}

\noindent
\textbf{Extending General Mask Transformers for PPS.} One simple way to extend Mask Transformers for the PPS task is to add part queries, as shown in Fig.~\ref{fig:meta_architecture}(a). Although it can serve as the simplest unified baseline, the limitation lies in the source of features for cross-reasoning. The part features are extremely local and fine-grained, while the thing/stuff features are mostly at global scopes. Thus, they naturally conflict with each other. Moreover, the part queries should be jointly modeled with the thing and stuff queries, so simply splitting the cross-reasoning on heads is not a good solution. 
We demonstrate the effectiveness of our meta-architecture in Sec.~\ref{sec:meta_architecutre_exp}.

\noindent
\textbf{Panoptic-PartFormer Meta-Architecture.} To handle these problems, we present Panoptic-PartFormer. 
First, we decouple both part features and thing/stuff features from the encoder. 
Then, the cross-attention is performed separately while the relationship is explored jointly. 
Finally, the decoder takes three different queries and their corresponding features as inputs and provides thing, stuff, and part mask predictions as outputs. 
Fig.~\ref{fig:meta_architecture}(b) shows the meta-architecture of our Panoptic-PartFormer, which will be described in Sec.~\ref{sec:ppformer_eccv} and Sec.~\ref{sec:ppformer++}.
Panoptic-PartFormer++ inherits the design of Panoptic-PartFormer as presented in Sec.~\ref{sec:ppformer_eccv} and  Sec.~\ref{sec:ppformer++}.

\begin{figure*}[t]
\centering
\includegraphics[width=0.8\linewidth]{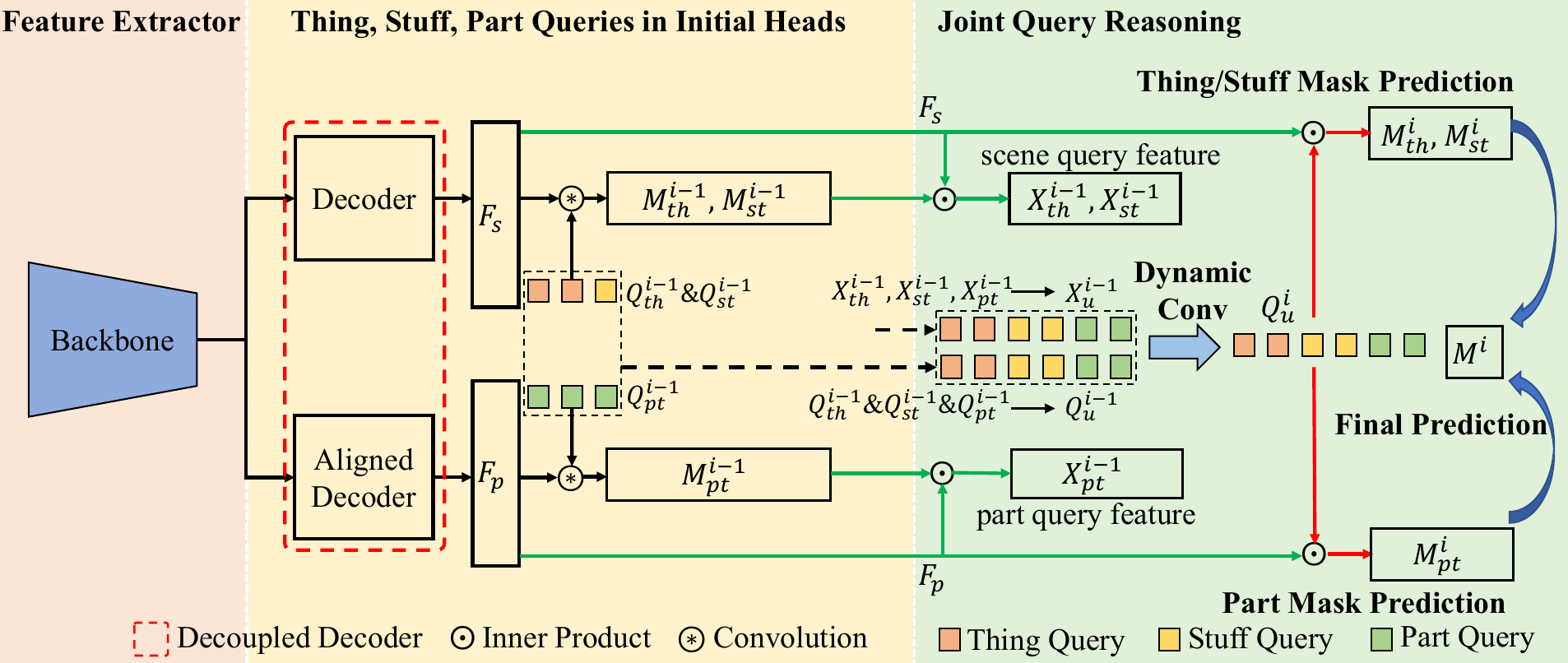}
\caption{Our proposed Panoptic-PartFormer. It contains three parts: (a) a backbone to extract features (Red area), (b) a decoupled decoder to generate scene features and part features along with the initial prediction heads to generate initial mask predictions. (Yellow area), (c) A cascaded transformer decoder will conduct the reasoning between the query and query features (Green area). \textcolor{green}{Green arrows} mean input (come from the previous stage) while \textcolor{red}{Red arrows} represent current stage output (used for the next stage). The outputs in the Red arrows are the inputs in the Green arrows. We take the last stage outputs as final output~\cite{detr,cheng2021mask2former}.}
\label{fig:methods}
\end{figure*}

\noindent
\textbf{Meta-Architecture Overview.} Fig.~\ref{fig:methods} presents the overall illustration of Panoptic-PartFormer as an example.
It contains three parts: (1) an encoder backbone to extract features; (2) a decoupled decoder to obtain the scene features and part features individually. Note that the \textit{scene features} are used to generate thing and stuff masks, while the \textit{part features} are used to generate part masks; (3) a Transformer decoder that takes three different types of queries and backbone features as inputs and provides thing, stuff, and part mask predictions. Both Panoptic-PartFormer and Panoptic-PartFormer++ share the \textbf{same} meta-architecture. We present the design details in Sec.~\ref{sec:ppformer_eccv} and Sec.~\ref{sec:ppformer++}.

\noindent
\textbf{Encoder.} We first extract image features using an encoder. It contains a backbone network (Convolution Network~\cite{resnet} or ViTs~\cite{liu2021swin}) with Feature Pyramid Network~\cite{fpn} as the neck. This results in a set of multiscale features, which play as the inputs of the decoupled decoder.

\noindent
\textbf{Decoupled Decoder.} The decoupled decoder has two separate decoder networks to obtain features for scene features and part features, respectively. 
The former decodes both thing and stuff predictions, while the latter is applied to decode the part prediction. 
Our motivation is that part segmentation has different properties from panoptic segmentation. 
Firstly, part features need a more precise location and fine details. 
Secondly, the scene features focus on mask proposal level prediction while part features pay more attention to the inner parts of mask proposal, which conflict with each other. 
We show that the decoupled design leads to better results in the experimental section (see Sec.~\ref{sec:ablation}). 
For implementation, we adopt semantic FPN design~\cite{kirillov2019panopticfpn} to fuse features in a top-down manner. 
Thus, we obtain the two features named $F_{s}$ and $F_{p}$. 
In particular, we design a lightweight aligned feature decoder for part segmentation. Rather than the naive bilinear upsampling, we propose to learn feature flow~\cite{sfnet,zhou2021differentiable} to warp the low-resolution feature to the high-resolution feature. 
Then, we sum all the predictions into the highest resolution as semantic FPN. Moreover, to preserve more locational information, we add the positional embedding to each stage of the semantic FPN, as adopted in the previous works~\cite{wang2020solo,wang2020solov2}.
As such, a decoupled decoder outputs two separated features: scene features $F_{s}$, and part features $F_{p}$. 
The former is used to generate thing and stuff masks, while the latter is used to generate part masks, and both have the exact resolution. 
We show the effectiveness of such design for both Panoptic-PartFormer and Panoptic-PartFormer++.

\subsection{Panoptic-PartFormer}
\label{sec:ppformer_eccv}

\noindent
\textbf{Thing, Stuff, and Part as Queries with Initial Head Prediction:} Previous works~\cite{cheng2021maskformer,zhang2021knet,wang2020maxDeeplab} show that single mask classification and prediction can achieve the state-of-the-art results on the COCO dataset~\cite{coco_dataset}. 
Motivated by this, our model treats thing, stuff, and part as the input queries to directly obtain the final panoptic part segmentation. 
Similar to prior works~\cite{zhang2021knet,peize2020sparse}, the initial weights of these queries are directly obtained from the first stage weights of the initial decoupled decoder prediction. 
For mask predictions of thing, stuff, and part, we use three $1 \times 1$ convolution layers to obtain the initial outputs of thing, stuff, and part masks. These layers are appended at the end of the decoupled decoder.

All these predictions are directly supervised with corresponding ground truth masks. 
As shown in~\cite{peize2020sparse,zhang2021knet}, using such initial heads can avoid heavier Transformer encoder layers for pixel-level computation, since the corresponding query features can be obtained via mask grouping from the initial mask prediction. 
Similarly, we obtain the three different queries for things, stuff, and parts, along with their initial mask predictions. 
We term them as $Q_{th}$, $Q_{st}$, $Q_{pt}$ and $M_{th}$, $M_{st}$, $M_{pt}$ with shapes $N_{th} \times d$, $N_{st} \times d$, $N_{pt} \times d$ and shapes $N_{th} \times H \times W$, $N_{st} \times H \times W$, $N_{pt} \times H \times W$. $d$, $W$, $H$ are the channel number, width, height of feature $F_{p}$ and $F_{s}$.  $N_{th}$, $N_{st}$, $N_{pt}$ are the numbers of queries for things, stuff, and parts. 

\noindent
\textbf{Joint Thing, Stuff, and Part Query Reasoning:} The cascaded Transformer decoder takes previous mask predictions, previous object queries, and decoupled features as inputs and outputs the current refined mask predictions and object queries. 
The refined mask predictions, object queries, and decoupled features will be the inputs of the next stage. 
The relationship between queries and query features is jointly learned and reasoned. 
The main ideas of our approach are stated as follows. 
First, joint learning can learn the complete correlation between scene features and part features. For example, car parts must be on the road rather than in the sky. Second, joint reasoning can avoid several scene noisy cases, such as car parts on the human body or human parts on the car. We find that joint learning leads to better results (see Sec.~\ref{sec:ablation}).

We combine the three queries and the three mask predictions into a unified query $Q_{u}^{i-1}$ and $M_{u}^{i-1}$ where $Q_{u}^{i-1} = \textrm{concat}(Q_{th}, Q_{st}, Q_{pt}) \in \boldsymbol{R}^{(N_{th} + N_{st} + N_{pt}) \times d}$ and $M_{u}^{i-1} = \textrm{concat}(M_{th}^{i-1}, M_{st}^{i-1}, M_{pt}^{i-1}) \in \boldsymbol{R}^{(N_{th} + N_{st} + N_{pt}) \times H \times W}$. 
Here, $i$ is the stage index of our Transformer decoder, and $i=1$ means the predictions come from the initial heads. 
Otherwise, it means the predictions come from the outputs of the previous stage.
In addition,  ``$\textrm{concat}$'' is the concatenating operation performed along the first dimension. 
Similar to~\cite{zhang2021knet}, we first obtain query features $X^i$ via grouping from previous mask predictions $M_{u}^{i-1}$ and input features ($F_{p}$, $F_{s}$) shown in Eq.~\ref{equ:grouping} (dot product in Fig.~\ref{fig:methods}). 
We present this process in one formulation for simplicity,
\begin{equation}
    X^i = \sum_{m=1}^W\sum_{n=1}^H M^{i-1}(m, n) \cdot F(m, n),
 \label{equ:grouping}
\end{equation}
where $X^i \in \boldsymbol{R}^{(N_{th} + N_{st} + N_{pt}) \times d}$ is the per-instance extracted feature with the same shape as  $Q_{u}$, $M^{i-1}$ is the per-instance mask extracted from the previous stage $i-1$, and $F$ is the input feature extracted for the decoupled decoder head. 
In addition, $m$, $n$ are the indices of spatial location, and  $i$ is the layer number and starts from 1. 
As shown in the center part of Fig.~\ref{fig:methods}, the part mask prediction and scene mask prediction are applied on corresponding features ($F_{p}$, $F_{s}$) individually where we obtain part query features $X_{pt}^i$ and scene query features $X_{th}^i$ and $X_{st}^i$. 
Then we combine these query features through $X_{u}^i = \textrm{concat}(X_{th}^i, X_{st}^i, X_{pt}^i)$. These inputs are shown in the green arrows in Fig.~\ref{fig:methods}.

We use dynamic convolution~\cite{zhang2021knet,peize2020sparse} to refine input queries $Q_{u}^{i-1}$ with features $X_{u}^i$ which are grouped from their masks,
\begin{equation}
    \hat{Q}_{u}^{i-1} = \mathrm{DynamicConv}(X_{u}^{i}, Q_{u}^{i-1}),
 \label{equ:dynamic}
\end{equation} where the dynamic convolution uses the query features $X_{u}^i$ to generate parameters to weight input queries $Q_{u}^{i-1}$. 
Specifically, $\mathrm{DynamicConv}$ uses input query features $X_{u}^{i}$ to generate gating parameters via MLP and multiply back to the original query input $Q_{u}^{i-1}$. 
The reason is two-fold. First, compared with pixel-wise MHSA~\cite{cheng2021maskformer,detr}, dynamic convolution introduces less computation and faster convergence. % for limited computation.
Second, it dynamically poses the instance-wise information to each query during training and inference, which shows better generalization and has complementary effects with MHSA. 
More details can be found in Sec.~\ref{sec:ablation}.

This operation absorbs more fine-grained information to help query and look for more precise locations. 
We adopt the same design~\cite{zhang2021knet} by learning gating functions to update the refined queries. 
The $\mathrm{DynamicConv}$ operation is based on:
\begin{equation}
    \hat{Q}_{u}^{i-1} = \mathrm{Gate}_{x}(X_{u}^{i})X_{u}^{i} + \mathrm{Gate}_{q}(X_{u}^{i}) Q_{u}^{i-1},
\end{equation}
where $\mathrm{Gate}$ is implemented with a fully connected (FC) layer followed by Layer Norm (LN) and a sigmoid layer, we adopt two different gate functions, including $\mathrm{Gate}_{x}$ and $\mathrm{Gate}_{q}$. 
The former is to weigh the query features, while the latter is to weigh corresponding queries. 
Next, we adopt one self-attention layer with feed-forward layers~\cite{vaswani2017attention,wang2020maxDeeplab} to learn the correspondence among each query and update them accordingly. 
This operation leads to the full correlation among queries, shown as follows:
\begin{equation}
    Q_{u}^{i} = \mathrm{FFN}(\mathrm{MHSA}(\hat{Q}_{u}^{i-1}) + \hat{Q}_{u}^{i-1}),
 \label{equ:selfattention}
\end{equation}
where $\mathrm{MHSA}$ means Multi Head Self Attention, $\mathrm{FFN}$ is the Feed Forward Network commonly used in the current vision of Transformers~\cite{detr,VIT}. 
The output refined query has the same shape as the input, \emph{i.e.}, $Q_{u}^{i} \in \boldsymbol{R}^{(N_{th} + N_{st} + N_{pt}) \times d}$. 

Finally, the refined masks are obtained via dot product between the refined queries $Q_{u}^i$, and the input features $F_{p}$, $F_{s}$. 
For mask classification, we adopt several feed-forward layers on $Q_{u}^{i}$ and directly output the class scores (thing, stuff, and part). 
For mask prediction, we also adopt several feed-forward layers on $Q_{u}^{i}$ and then we perform the inner product between the learned queries and features ($F_{s}$ $F_{p}$) to generate scene masks (thing and stuff) and part masks of stage $i$.
These masks will be used for the following stage input, as shown in the red arrows in Fig.~\ref{fig:methods}. 
The process of Eq.~\ref{equ:grouping}, Eq.~\ref{equ:dynamic}, and Eq.~\ref{equ:selfattention} are repeated several times.
We set the iteration number to 3 in this work.
The inter-mask predictions are also optimized by mask supervision.

\subsection{Part-Whole Quality Metric}
\label{sec:PWQ}

\begin{table}[t]
\centering
\caption{Upper Bound Analysis using Merge baseline and PanopticPartFormer on the CPP dataset. We report the results using ResNet-50 Backbone. Our new proposed PWQ is not sensitive to the quality of panoptic segmentation, which is more balanced.}
\begin{adjustbox}{width=0.45\textwidth}
\begin{tabular}{c c c c c c} 
  \toprule[0.15em]
    Setting & Panoptic-GT  & Part-GT  & PartPQ & PartPQ$_{part}$ & PWQ \\
    \toprule[0.15em]
       Merged baseline & - & - & 56.8 & 42.5 & 60.8 \\
        & - & \checkmark & 60.8 & 55.9 & 75.2 \\
        & \checkmark &  - & 85.4 & 54.4 & 78.3 \\
        \hline
        Panoptic-PartFormer  & - & - & 57.4 & 43.9 & 62.1 \\
        & - & \checkmark & 61.6 & 56.1 & 76.9 \\
        & \checkmark &  - & 88.4 & 56.4 & 79.8 \\
    \bottomrule[0.1em]
\end{tabular}
\end{adjustbox}
\label{tab:upper_bound_analysis}
\end{table}

\begin{table}[t]
\centering
\caption{PPS Metric Analysis. The proposed PWQ has all five important properties.}
\begin{adjustbox}{width=0.45\textwidth}
\begin{tabular}{c c c c c c} 
  \toprule[0.15em]
    Metric Properties & mIOU & PQ  &  PartPQ & HPQ & PWQ  \\
    \toprule[0.15em]
    Pixel-Level Evaluation  & \checkmark & - & - & - &  \checkmark\\
    Region-Level Evaluation  & - & \checkmark &  \checkmark & \checkmark &\checkmark \\   
    Part-Whole Evaluation & - & - & \checkmark & \checkmark & \checkmark \\
    Decouple Errors & - & - & - & - & \checkmark \\
    Balance Part and Scene Segments & - & -  & - & - & \checkmark\\
    \bottomrule[0.1em]
\end{tabular}
\end{adjustbox}
\label{tab:metric_analysis}
\end{table}

\noindent
\textbf{Issues of PartPQ.} There are several issues with the PartPQ metric, as defined in Eq.~\ref{equ:partPQ}. 
First, PartPQ biases against PQ. 
We use two different models: Panoptic-ParFormer and Merged baseline~\cite{zhang2021knet} to analyze the upper bound performance by replacing the panoptic segmentation Ground Truth (GT) or part segmentation GT into our prediction. 
The Merged baseline contains a panoptic segmentation model~\cite{zhang2021knet} and one separate part segmentation model~\cite{deeplabv3plus}. For both models, we find that replacing panoptic segmentation ground truth leads to a significant gain on PartPQ, while replacing part segmentation only results in a limited gain. 
This indicates that PartPQ is more \textit{sensitive to panoptic segmentation than part segmentation} on the CPP dataset, which decreases the role of part segmentation. 
Second, it only considers region-level matching and lacks pixel-level evaluation.
Third, it does not decouple the error of part and scene segmentation, which lacks interpretability of model performance. 
Namely, the PartPQ metric does not balance the two types of relations: part-whole relation and pixel-region relation. 

\noindent
\textbf{Proposed Part-Whole Quality.} Motivated by the previous analysis, we present a new metric named Part-Whole Quality. 
The main idea is to decouple the error of part and scene segmentation outputs and consider both region-level and pixel-level evaluation in one formulation. Our metric is defined as:
\begin{equation}
\textrm{PWQ} =  \left(\frac{\textrm{mIoU}_{th, st} \cdot \textrm{PQ}_{th, st} + \textrm{mIoU}_{part} \cdot \textrm{PartPQ}_{part}}{2}\right)^\frac{1}{2},
 \label{equ:PWQ_quality}
\end{equation}
where $\textrm{PartPQ}_{part}$ only considers the $\textrm{PartPQ}$ metric for part segments, and $\textrm{mIoU}$ refers to the mean Intersection over Union for scene segments (including thing (th) and stuff (st) masks) and part segments. 
In particular, we adopt an average of scene and part to highlight the ratio of part segments. The $\textrm{PWQ}$ metric can be decoupled into two items: Scene Segment Quality: $\textrm{SSQ}= \textrm{mIoU}_{th, st} \cdot \textrm{PQ}_{th, st}$ and Part Segment Quality: $\textrm{PSQ} = \textrm{mIoU}_{part} \cdot \textrm{PartPQ}_{part}$. The $\textrm{PWQ}$ is defined as the geometric mean of $\textrm{SSQ}$ and $\textrm{PSQ}$. 
As shown in the last column of Tab.~\ref{tab:metric_analysis}, compared with previous metrics, PartPQ and HPQ, our new metric fully considers all five properties, including pixel-level evaluation, region-level evaluation, part-whole evaluation, decoupling the errors, and balancing the part-scene segments. 
We note that the recent work~\cite{tang2022visual} presents a hierarchical panoptic quality (HPQ) metric to measure the accuracy of segmentation in different depths. 
However, it still does not well balance the ratio of part segments.
After using the proposed PWQ, as shown in the last column of Tab.~\ref{tab:upper_bound_analysis}, by the same upper bound analysis, the proposed PWQ is more friendly to balancing the part segments and scene segments. 

\begin{figure}[!t]
	\centering
        \includegraphics[width=0.90\linewidth]{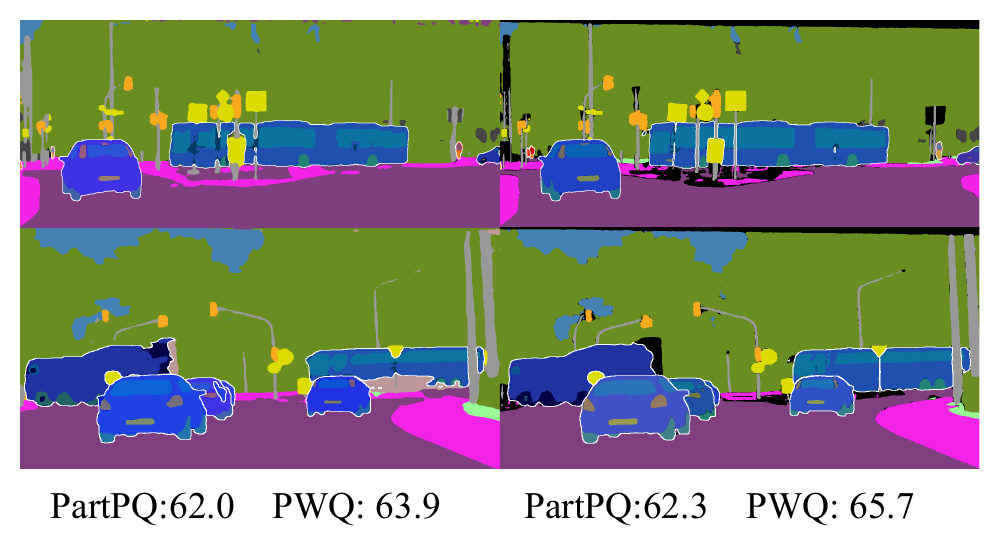}
	\caption{Visual examples for PartPQ and PWQ. Left: Merged baseline. Right: Our PanopticPartFormer++. Both methods use Swin-base as the backbone. Although both methods achieve similar PartPQ results. The part segmentation quality is significantly different. Our proposed PWQ indicates the better difference between both methods for part details.}
    \label{fig:visual_issues_partPQ}
\end{figure}

We present several cropped examples in Fig.~\ref{fig:visual_issues_partPQ}. We use a stronger merged baseline~\cite{degeus2021panopticparts} for illustration of the visual issues of PartPQ. 
The left-hand side of this figure is the same merge baseline as adopted in Tab.~\ref{tab:upper_bound_analysis}, and the right-hand side is the proposed Panoptic-PartFormer++, which will be detailed further. 
Both methods use the same Swin-base backbone. Although both methods achieve similar PartPQ results, it is shown from Fig.~\ref{fig:visual_issues_partPQ} that our proposed PWQ shows a clearer difference between the two methods, which is in line with the contrast of visual qualities.

\begin{figure*}[t]
\centering
\includegraphics[width=0.8\linewidth]{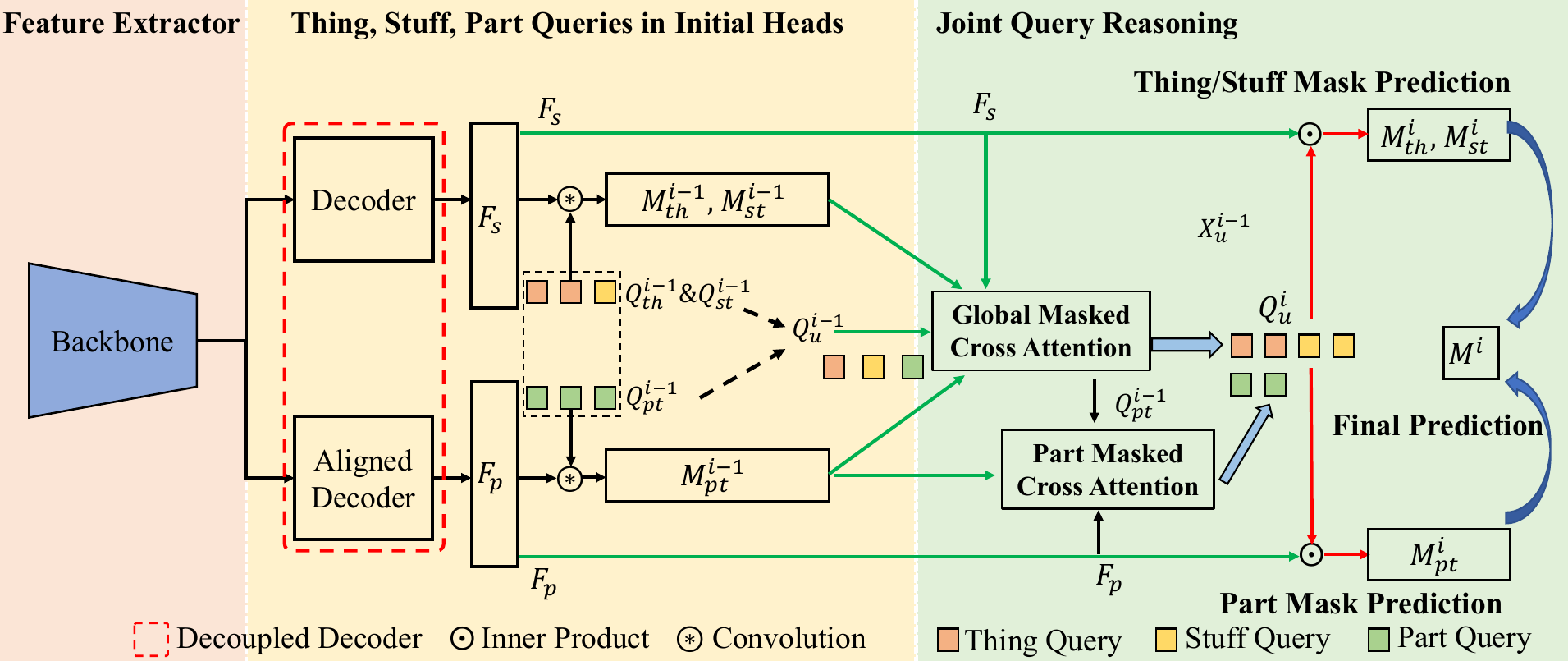}
\caption{Proposed Panoptic-PartFormer++. The Panoptic-PartFormer++ follows the meta-architecture design of Panoptic-PartFormer. It proposes a new Global-Part Masked Cross Attention design by performing joint query reasoning and then doing cross-attention on local part features. Global Masked Cross (GMC) attention takes object queries, object masks, and scene features $F_{s}$ as inputs and outputs the refined object queries and scene masks. Part Masked Cross (PMC) attention takes the refined part queries from GMC, and part features $F_{p}$ as inputs and outputs the refine part queries and part masks. Best viewed in color and zoom in.}
\label{fig:method_panopart_plus}
\end{figure*}

\subsection{PanopticFormer++}
\label{sec:ppformer++}

\noindent
\textbf{Motivation.} As shown in the analysis of Sec.~\ref{sec:PWQ}, to obtain a better PPS model, both scene-level and part-level segment outputs should be improved. In Panoptic-Partformer, 
we argue that three different issues still exist. \textbf{First}, the masked grouping operation (Eq.~\ref{equ:grouping}) misses the fine-grained details for both scene and part features. 
\textbf{Secondly}, although local part queries should guided by high-level scene queries, the local details are still important, and they should not be dominated by scene features.
\textbf{Thirdly}, a stronger feature extractor can lead to both scene-level and part-level segmentation, which is still not explored.
Solving these issues can improve the PSS performance for various metrics.

%
% We propose a new design named Global-Part Masked Cross Attention for joint query reasoning to better explore the relationship between global scene features and part features. The entire pipeline is shown in Fig.~\ref{fig:method_panopart_plus}.

\noindent
\textbf{Enhanced Feature Extractor.} 
For the last issues, we integrate our Meta-Architecture in Fig.~\ref{fig:meta_architecture} with the Mask2Former design. We show that a better PPS model can improve both scene and part masks by using Mask2Former~\cite{cheng2021mask2former} as the baseline model for feature extraction and fine-grained cross-reasoning. 
We use the deformable Feature Pyramid Networks~\cite{zhu2020deformabledetr} (deformable FPN) in Mask2Former to replace the feature extractor of Panoptic-PartFormer, and retain the
decoupled decoder. 
The aligned decoder is appended after the output of deformable FPN.
Similar to~\cite{cheng2021mask2former}, we also adopt the positional embedding on both scene and part features. 

%HERE
\noindent
\textbf{Revisiting Joint Query Reasoning Design.} The PanopticPartFormer uses a joint query reasoning design to simplify the PPS pipeline. 
However, as shown in Sec.~\ref{sec:PWQ}, due to the unbalanced issues of PartPQ, previous design~\cite{li2022panopticpartformer} overlooks the key role of part segmentation design. 
To enhance the part segmentation quality, we present a compelling scene and part query reasoning method by leveraging the benefits of joint global reasoning and local part features, \emph{i.e.}, global-part masked cross attention. 

\noindent
\textbf{Global-Part Masked Cross Attention.} This design mainly solves the first two issues. This process contains two steps: First, we perform joint query reasoning by extending masked cross attention on the three object queries and scene features. (\textbf{Global Masked Cross attention})
Second, we use refined part queries to perform masked cross-attention on part features. (\textbf{Part Masked Cross attention})

As shown in the experiment results of Panoptic-PartFormer in Sec.~\ref{sec:ablation}, joint learning with scene features and part features leads to better part segmentation. 
However, adding part supervision leads to a minor effect on scene segmentation. 
Thus, a better joint query reasoning design should be well-designed. 
We analyze the masked cross attention:
\begin{align}
    \label{equ:masked_cs_attention}
    {Q}_{l} = \mathrm{softmax}({M}_{l-1} + \mathrm{MLP}({Q}_{l}){K}_{l}^{T}){V}_{l} + {Q}_{l-1},
\end{align}
where ${M}_{l-1}$ is a 2D attention mask from the previous stage $l$,  ${M}_{l-1} \in \mathbb{R}^{n \times H\times {W}} $, and 
$\mathrm{MLP}$ is a Multilayer Perceptron network that contains two fully connected layers. 

As Global Masked Cross (GMC) attention mainly models the global relation of scene and part queries, we combine all mask predictions into $M_{u}^{i-1}$ and corresponding queries $Q_{u}^{i-1}$.
We perform joint attention and reasoning as in Eq.~\ref{equ:dynamic} and Eq.~\ref{equ:selfattention}.
We first replace the $Q_{l}$ with $Q_{u}^{i-1}$, $M_{l-1}$ with $M_{u}^{i-1}$ in Eq.~\ref{equ:masked_cs_attention}. $K_{l}$ and $V_{l}$ are obtained via different MLPs on shared scene feature $F_{s}$.
The operation works similarly to the previous dynamic convolution step in Eq.~\ref{equ:dynamic}. 
However, previous pooling-based approaches~\cite{zhang2021knet,li2022panopticpartformer} miss the detailed information, which is more important for part queries.
Using pixel-level cross attention injects more fine-grained information into all three queries, which leads to better results (Sec.~\ref{sec:exp_main_results}).
Then, we perform the self-attention and FFN on refined $Q_{u}$ to learn the global correction of thing, stuff, and part query.
This operation works the same as Eq.~\ref{equ:selfattention}.
For brevity, we ignore this operation for notation.
Although Mask2Former contains multiscale features, for notation brevity, we use the feature with the highest mask resolution $F_{s}$ for formulation purposes. 
The scene masks are classified with output queries. The thing/stuff mask predictions are obtained via inner production with $F_{s}$. 
For scene features, we adopt the multiscale features to produce the thing/stuff masks. Via pixel-level cross attention, we can achieve stronger results on both scene segmentation and part segmentation.

Then, for the second issue, we further present a refined step to inject more local features into part queries. In particular, as for Part Masked Cross (PMC) attention, we take part queries $Q_{pt'}^{i-1}$ from GMC attention outputs and corresponding part masks $M_{pt}^{i-1}$ to replace $Q_{l}$ and $M_{l-1}$ in Eq.~\ref{equ:masked_cs_attention}. 
In this module, $K_{l}$ and $V_{l}$ are obtained via MLP on part feature $F_{p}$, and 
$Q_{pt'}^{i-1}$ has already absorbed the global context information and can be further refined by the local fine detailed feature $F_{p}$.
The part masks are classified with final part queries. The part mask predictions are obtained via inner production with $F_{p}$. 
For both global and part masked cross attention, we perform MHSA operation in Eq.~\ref{equ:selfattention} after the cross attention. 
Unlike the Global Masked Cross Attention performed multiple times on different resolution features, the Part Masked Cross Attention is only performed on the \textit{highest resolution} feature. 
In our implementation, we add an extra dense prediction head to force the part features sensitive to the part structure. We use the same Mask2Former design, \emph{i.e.}, both Masked Cross Attentions are repeated multiple times. 
Moreover, after each cross attention in Eq.~\ref{equ:masked_cs_attention}, we perform self-attention and feed-forward networks on output queries, which is similar to Eq.~\ref{equ:selfattention}.

\noindent
\textbf{Discussion on Cross Attention Design.} 
%We admit that we use the dynamic convolution, masked cross-attention, and self-attention among queries proposed by~\cite{peize2020sparse,zhang2021knet,cheng2021mask2former}. However, we \textbf{\textit{do not claim}} this is our core contribution for both Panoptic-PartFormer and Panoptic-PartFormer++. Our \textbf{\textit{main contribution}} is a \textbf{system-level unified model} for this challenging task (PPS). 
While our model uses dynamic convolution, masked cross-attention, and self-attention among queries, 
the main contribution of this work is the system-level unified model for this challenging task (PPS).
We show that joint learning of the thing, stuff, and part learning benefits PPS tasks more than many other designs. 
In addition, we demonstrate that learning joint global relations first and then performing part segmentation is a good design for part segmentation. 
Furthermore, we verify these designs in the experiment section (Sec.~\ref{sec:ablation_pppformer_plus}).

%We also present a detailed comparison in the appendix file.

\subsection{Training and inference} 
\label{sec:train_and_inference}

\noindent
\textbf{Training.} To train Panoptic-PartFormer and Panoptic-PartFormer++, we need to assign ground truth according to the pre-defined cost since all the outputs are encoded via queries. 
Similar to~\cite{cheng2021maskformer}, we use bipartite matching as a cost by considering both mask and classification results. 
After the bipartite matching, we apply a loss jointly considering mask prediction and classification for each thing, stuff, and part. 
We apply cross-entropy loss on both classification and mask prediction. 
In addition, we adopt dice loss~\cite{dice_loss} on mask predictions ($L_{part}$, $L_{thing}$, $L_{stuff}$). Such settings are \textit{the same as previous works}~\cite{detr,cheng2021maskformer}. 
The loss for each stage $i$ is: 
\begin{equation}
    \mathcal{L}_{i} = \lambda_{part} \cdot \mathcal{L}_{part} + \lambda_{thing} \cdot \mathcal{L}_{thing} + \lambda_{stuff} \cdot \mathcal{L}_{stuff} +\lambda_{cls} \cdot \mathcal{L}_{cls}
\end{equation}
Note that the losses are applied to each stage, $\mathcal{L}_{final} = \sum_{i}^N\mathcal{L}_i,$ where $N$ is the total stages applied to the framework. 
Moreover, for Panoptic-PartFormer++, we add an extra semantic segmentation loss on part segmentation with part feature $F_{p}$.

\noindent
\textbf{Inference:} We directly get the output masks from the corresponding queries according to their sorted scores. To obtain the final panoptic part segmentation, we first obtain the panoptic segmentation results and then merge part masks into panoptic segmentation results. 
For panoptic segmentation, we adopt the method in Panoptic-FPN~\cite{kirillov2019panopticfpn} to merge the panoptic mask. 
%
%We use a process similar to the PPS task to merge parts for the final panoptic part segmentation results. 
%
We copy the predictions from panoptic segmentation for scene-level semantic classes that do not have part classes. 
For predicted instances with parts, we extract the part predictions for the pixels corresponding to this segment. 
Otherwise, if a part prediction contains a part class that does not correspond to the scene-level class, we set it to \textit{void} label. 
%
%This setting is similar to previous work~\cite{degeus2021panopticparts}.

	\section{Experiments}
% Setting 
% Results
% Ablations
% Visualization and Analysis

\subsection{Experiment Setup}
\label{sec:exp_set_up}

\noindent
\textbf{Datasets.} We carry out experiments on the Cityscapes Panoptic Parts (CPP) and PASCAL Panoptic Parts (PPP) datasets.
%which are based on the established scene understanding datasets Cityscapes~\cite{cordts2016cityscapes} and PASCAL VOC~\cite{Everingham2010Pascal}, respectively. 
%
The CPP dataset is constructed based on the Cityscapes set~\cite{cordts2016cityscapes} with part-level semantics and is annotated with 23 part-level semantic classes. 
Five scene-level semantic classes from the {human} and {vehicle} high-level categories are annotated with parts. 
The CPP dataset contains 2,975 training and 500 validation images.
The  PPP dataset extends the PASCAL VOC 2010 benchmark~\cite{Everingham2010Pascal} with part-level and scene-level semantics. PPP has 4,998 training and 5,105 validation images. 
For fair comparisons, we use the same settings as~\cite{degeus2021panopticparts,Everingham2010Pascal}, for experiments with 59 scene-level classes (20 things, 39 stuff) and 57 part classes. 
We further report the Cityscapes Panoptic Segmentation validation set~\cite{cordts2016cityscapes} results for performance evaluation. 

%HERE
\noindent
\textbf{Implementation Details.} In the proposed method, ResNet~\cite{resnet} and Vision Transformer~\cite{liu2021swin,bao2021beit} are adopted as the backbone networks, and other layers use Xavier initialization~\cite{xavier_init}. 
The optimizer is AdamW~\cite{ADAMW} with a weight decay of 0.0001. 
The training batch size is set to 16, and all models are trained with 8 GPUs. 
For the PPP dataset, we first pre-train our model on the COCO dataset~\cite{coco_dataset} since most previous baselines~\cite{degeus2021panopticparts} are also pre-trained on the COCO dataset. 
For the PPP dataset, we adopt the multiscale training~\cite{detr} by resizing the input images from a scale of 0.5 to a scale of 2.0.
We also apply random crop augmentations during training, where the training images are cropped with a probability of 0.5. Each random rectangular patch is then resized again to 800 (height), 1,333 (width) for training.
For the CPP dataset, we use the same setting in Panoptic-Deeplab~\cite{cheng2020panoptic} where we resize the images with a scale range from 0.5 to 2.0 and randomly crop the whole image during training with a batch size of 8. 
We adopt V100 GPU servers with 16 32GB GPUs for training and testing. 
All the results are obtained via single-scale inference. We refer to the COCO-pretraining and Mapillary-pretraining for large models in the appendix.

\noindent
\textbf{Evaluation Metric.} 
Aside from the PartPQ and PQ measures, we use our proposed PWQ metric to analyze the evaluated methods.
For PanopticPart-Former, we adopt the PartPQ and PQ metrics for analysis. 
For PanopticPart-Former++, we use the PartPQ and PWQ measures (including SSQ and PSQ) for analysis.

\begin{table*}[t]
\centering
\caption{\textbf{Experiment Results on the CPP dataset.} 
Prior works~\cite{degeus2021panopticparts,tang2022visual} combine results from commonly used (top), and state-of-the-art methods (bottom) for semantic segmentation, instance segmentation, panoptic segmentation, and part segmentation. 
Metrics split into \textit{P} and \textit{NP} are evaluated on scene-level classes with and without parts, respectively. 
Our proposed Panoptic-PartFormer and Panoptic-PartFormer++ provide a simple yet unified solution with better performance and much fewer GFlops. We list the improvements of PanopticPartFormer++ by \textcolor{black}{black $\uparrow$}.}
\begin{adjustbox}{width=0.85\textwidth}
%\begin{tabular}{ll||ccc||ccc||cccc}
\begin{tabular}{llcccccccccc}
\toprule[0.15em]
 &  & \multicolumn{3}{c}{\textbf{PQ}} & \multicolumn{3}{c}{\textbf{PartPQ}} & \textbf{PWQ} \\ 
\textbf{Panoptic seg. method}  &  \textbf{Part seg. method}  & All & P & NP & All & P & NP \\ 
\toprule[0.15em]
 
\textit{Settings: \textbf{Cityscapes Panoptic Parts} validation set} \\
UPSNet \cite{xiong2019upsnet}(ResNet50) & DeepLabv3+ \cite{deeplabv3plus}(ResNet50)  & 59.1 & 57.3 & 59.7  & 55.1 & 42.3 & 59.7 & -\\
DeepLabv3+(ResNet50) \& Mask R-CNN(ResNet50) \cite{maskrcnn} & DeepLabv3+ \cite{deeplabv3plus} (Xception-
65)  & 61.0 & 58.7 & 61.9 &  56.9 & 43.0 & 61.9 & -  \\
SegFormer-B1 \& CondinInst~\cite{tang2022visual} & SegFormer-B1~\cite{xie2021segformer} & - & - & - & 57.9 & 47.0 & 61.9 & - \\ 
\hline
\textbf{Panoptic-PartFormer (ResNet50)}~\cite{li2022panopticpartformer}  & None & {61.6} & {60.0} & {62.2} & {57.4} & {43.9} & {62.2} &  60.5 \\ 
\textbf{Panoptic-PartFormer++ (ResNet50)}  & None & {63.6} (\textcolor{black}{2.0 $\uparrow$}) & {61.0} & {63.1} & {59.2}(\textcolor{black}{1.8 $\uparrow$}) & {42.5} & {65.1} & 62.1 (\textcolor{black}{1.6 $\uparrow$})\\ 
\hline

EfficientPS \cite{mohan2021efficientps}(EfficientNet)~\cite{tan2019efficientnet} &  BSANet \cite{Zhao2019BSANet}(ResNet101)   & 65.0 & 64.2 & 65.2   & 60.2 & {46.1} & 65.2 & - \\
HRNet-OCR (HRNetv2-W48)~\cite{ocrnet,wang2020deep} \& PolyTransform~\cite{liang2020polytransform} &  BSANet \cite{Zhao2019BSANet}(ResNet101)  & 66.2 & 64.2 & 67.0  & 61.4 & 45.8 & 67.0 & -\\
SegFormer-B5 \& CondinInst~\cite{tang2022visual} & SegFormer-B5~\cite{xie2021segformer} & - & - & - & 61.2 & 48.1  & 65.8 & - \\ 
Part Based SegFormer-B5 \& CondinInst~\cite{tang2022visual} & SegFormer-B5~\cite{xie2021segformer} & - & - & - & 62.4 & 50.5  & 66.6  & - \\ 
\hline
\textbf{Panoptic-PartFormer (Swin-base)}~\cite{li2022panopticpartformer} & None & {66.6}  & {65.1}  & {67.2}  & {61.9}   & {45.6} & {68.0} & 64.4 \\
\textbf{Panoptic-PartFormer++ (Swin-base)} & None & 68.0 (\textcolor{black}{1.4 $\uparrow$}) & 67.0  & 68.3 & 62.3 (\textcolor{black}{0.4 $\uparrow$})  &  46.0 & 68.2  & 65.7 (\textcolor{black}{1.7 $\uparrow$})\\
\textbf{Panoptic-PartFormer++ (Convnext-base)} & None & 68.2 & 67.2 & 69.0    & {63.1} & 46.4 &  69.1 & 66.5 \\
\bottomrule
\end{tabular}
\end{adjustbox}
\label{tab:experiments_res_city}
\end{table*}

\begin{table}[t]
\centering
\caption{\textbf{Experiment Results on the PPP dataset.} Our methods using a \textit{smaller} backbone ResNet101 achieve new state-of-the-art results with much lower GFLops. Both models show the effectiveness of the end-to-end design. We list the improvements of PanopticPartFormer++ by \textcolor{black}{$\uparrow$}.}
\begin{adjustbox}{width=0.50\textwidth}
%\begin{tabular}{ll||ccc}
\begin{tabular}{llccc}
\toprule[0.15em]
\textbf{Panoptic seg. method}  &  \textbf{Part seg. method}  & PQ & PartPQ & PWQ \\ 
\midrule
\textit{Settings: \textbf{Pascal Panoptic Parts} validation set} \\
DeepLabv3+ \& Mask R-CNN \cite{maskrcnn}(ResNet50) & DeepLabv3+ \cite{deeplabv3plus}(ResNet50) & 35.0  &  31.4 & - \\ 
DLv3-ResNeSt269~\cite{zhang2020resnest} \& DetectoRS \cite{qiao2021detectors} &  BSANet \cite{Zhao2019BSANet}  & 42.0  & 38.3 & -  \\ \hline
\textbf{Our Unified Approach} \\
\textbf{Panoptic-PartFormer (ResNet50)} & None & {47.6} & {43.2} & 42.3 \\
\textbf{Panoptic-PartFormer (ResNet101)} & None & {49.2} & {44.0} & 43.6 \\
\hline
% 42.2
% 42.4
\textbf{Panoptic-PartFormer++ (ResNet50)} & None &  51.6 (\textcolor{black}{1.9 $\uparrow$}) &  45.1 (\textcolor{black}{1.8 $\uparrow$})  & 45.2 (\textcolor{black}{2.0 $\uparrow$}) \\
\textbf{Panoptic-PartFormer++ (ResNet101)} & None &  52.5 (\textcolor{black}{3.3 $\uparrow$}) &  45.6 (\textcolor{black}{1.9 $\uparrow$}) & 46.0 (\textcolor{black}{2.4 $\uparrow$}) \\
\bottomrule
\end{tabular}
\end{adjustbox}
\label{tab:experiments_res_ppp}
\end{table}

\begin{table}[!t]\setlength{\tabcolsep}{6pt}
	\centering
 	\caption{\textbf{GFlops and parameter comparison on the Cityscapes PPS dataset.} The GFlops are measured with 1200 $\times$ 800 input. Both proposed Panoptic-PartFormer and Panoptic-PartFormer++ achieve better results with much fewer GFlops and parameters.}
	\begin{adjustbox}{width=0.50\textwidth}
	\begin{tabular}{l c c c c}
				\toprule[0.15em]
			Method &  PQ & PartPQ  & Param(M) & GFlops \\
				\toprule[0.15em]
		UPSNet + DeepLabv3+ (ResNet50) & 59.1 & 55.1 & $>$87 &  $>$890 \\ 
		\textbf{Panoptic-PartFormer (ResNet50)} & {61.6} & {57.4}  & {37.4} & {185.8}  \\
        \textbf{Panoptic-PartFormer++ (ResNet50)} & {62.8} & {59.2}  & {45.6} & {215.4}  
            \\ 
		\hline
		HRNet(OCR) +PolyTransform + BSANet & 66.2 & 61.4 & $>$181 & $>$1154 \\
        \textbf{Panoptic-PartFormer (Swin-base)} & {66.6}  & {61.9}  & {100.3} & {408.5} \\
        \textbf{Panoptic-PartFormer++ (Convnext-base)} & {68.2}  & {63.1}  & {120.2} & {519.5} \\
	\bottomrule[0.1em]
	\end{tabular}
	\label{tab:gflops_parameter}
	\end{adjustbox}
\end{table}

\begin{table}[!t]\setlength{\tabcolsep}{6pt}
	\centering
 	\caption{\textbf{Scaling Up the Panoptic-PartFormer++}. We adopt the recent state-of-the-art backbones to explore the effectiveness of representation learning on PPS tasks. All the methods are pre-trained on COCO.}
          \label{tab:exp_scale_up}
	\begin{adjustbox}{width=0.40\textwidth}
	\begin{tabular}{l c c c c}
	\toprule[0.15em]
	Dataset &  Backbone & PQ & PartPQ & PWQ   \\
	\toprule[0.15em]
	CPP & Swin-base & 68.0 & 62.3 & 65.7 \\
        CPP & Convnext-base & 68.2 & 63.1 & 66.5 \\
        CPP & Convnext-large & 67.8  & 62.8 & 65.4\\
        CPP & BEIT-V2 & 67.5 & 62.7 & 66.0 \\
        \hline
        % 49.3 48.6 48.8
        PPP & Swin-base & 59.8 & 51.3 & 52.7 \\ 
        PPP & Convnext-base & 58.7 & 52.6 & 54.2 \\ 
        PPP & Convnext-large & 60.2 & 52.8 & 54.7  \\ 
	\bottomrule[0.1em]
	\end{tabular}
	\label{tab:scale_up}
	\end{adjustbox}
\end{table}

%HERE
\subsection{Main Results}
\label{sec:exp_main_results}

\noindent
\textbf{Results on Cityscape Panoptic Part Dataset.} Tab.~\ref{tab:experiments_res_city} shows the results of our Panoptic-PartFormer and other baselines. 
All the models use single-scale inference without test time augmentation. 
Our method with ResNe50 backbone achieves 57.4\% PartPQ, which outperforms the previous work using complex pipelines~\cite{deeplabv3plus,maskrcnn} with even stronger backbone~\cite{chollet2017xception}. 
For the same backbone, our method achieves \textbf{2.3\% PartPQ gain} over the previous baseline. 
For the large model comparison, our method with Swin-Transformer achieves 61.9 \% PartPQ. 
It outperforms the previous works that use state-of-the-art individual models~\cite{ocrnet,wang2020deep,Zhao2019BSANet,liang2020polytransform} by 0.5\%. 
Note that the best model from HRNet~\cite{ocrnet} is pre-trained using the Mapillary dataset~\cite{neuhold2017mapillary}. 
We use the same pipeline for fair comparisons. 
For our new proposed Panoptic-PartFormer++, we \textit{only} use the COCO dataset for pertaining. 
As shown in Tab.~\ref{tab:experiments_res_city}, the proposed Panoptic-PartFormer++ achieves about \textbf{1.3\%-1.6\% PartPQ and 1.6\% PWQ gain} over Panoptic-PartFormer. 
With the recently proposed backbone~\cite{liu2022convnet}, our method obtains the new state-of-the-art results with 63.1\% PartPQ and 66.5\% PWQ. 
Compared to the recent method using separated vision transformers~\cite{tang2022visual}, our methods perform favorably with both ResNet50 and a larger backbone with fewer GFlops and a more direct pipeline. 
In conclusion, compared to Panoptic-PartFormer, Panoptic-PartFormer++ achieves better results on all three metrics, including PQ, PartPQ, and PWQ, which can be a new baseline for the PPS task.

\begin{table*}[!h]
     \centering
    \caption{\textbf{Ablation studies and analysis on Panoptic-PartFormer using the CPP validation set.} All the models use ResNet50 as the backbone, with 64 training epochs. DD: Decoupled Decoder. DC: Dynamic Convolution. SA: Self Attention. I: Interaction number.DP: Dense Prediction. w: with. ASPP: Atrous Spatial Pyramid Pooling~\cite{deeplabv3}.  }
    \label{tab:ablation_ppformer}
    \subfloat[Effect of each component.]{
        \small
        \label{tab:effect_component}
	    \begin{tabularx}{0.35\textwidth}{c c c c c c c} 
		        				\toprule[0.15em]
    	DD & DC  & SA & I=1 & I=3 & PQ & PartPQ  \\
            \midrule[0.15em]
    	\rowcolor{gray!15}	\checkmark & \checkmark & \checkmark &  - & \checkmark & 61.6 & 57.4 \\
    		- & \checkmark & \checkmark & - & \checkmark & 61.2 & 55.9 \\ 
    		\checkmark & - & \checkmark & - & \checkmark & 57.0 & 52.2 \\ 
    		\checkmark & \checkmark & - &- & \checkmark & 57.3 & 53.4 \\ 
    		\checkmark & \checkmark & \checkmark &\checkmark & - & 58.3 & 54.2 \\ 
        	\bottomrule[0.1em]
	    \end{tabularx}
    } \hfill
    \subfloat[Ablation on Query Reasoning Design]{
        \label{tab:query_reasoning}
		\begin{tabularx}{0.30\textwidth}{c c c} 
			\toprule[0.15em]
		  Setting & PQ & PartPQ \\
			\midrule[0.15em]
		\rowcolor{gray!15}  Joint Reasoning & 61.6 & 57.4 \\
          Separate Reasoning & 61.1 & 56.8 \\
          Sequential Reasoning & 60.8 & 56.3 \\
			\bottomrule[0.1em]
		\end{tabularx}
    } \hfill
    \subfloat[Dense Prediction or Query Prediction on Part.]{
        \label{tab:dp_vs_jq}
		\begin{tabularx}{0.30\textwidth}{c c c} 
			\toprule[0.20em]
			Method & PQ & PartPQ \\
			\midrule[0.15em]
		\rowcolor{gray!15}	Joint Query  & 61.6 & 57.4 \\ 
            DP-Based & 59.8 & 55.9 \\ 
            DP-Based w ASPP~\cite{deeplabv3} & 59.9 & 56.1 \\ 
			\bottomrule[0.1em]
		\end{tabularx}
    } \hfill
    \subfloat[Effect of Aligned Decoder Design.]{
        \label{tab:aligned_decoder}
	    \begin{tabularx}{0.35\textwidth}{c c c c c} 
		        				\toprule[0.15em]
    		 Settings  & PQ & PartPQ & P & NP \\
    		\toprule[0.15em]
    	\rowcolor{gray!15}	w Aligned & 61.6 & 57.4 & 43.9 & 62.2\\
    		w/o Aligned & 61.4 & 56.3 & 41.2 & 62.1  \\
    		on Both Features &  61.4 & 57.2 & 43.7 & 62.0 \\
        	\bottomrule[0.1em]
	    \end{tabularx}
    }\hfill
    \subfloat[Effect of Position Encoding (PE) on $F_{s}$ and $F_{p}$.]{
     \label{tab:pos_enc}
	    \begin{tabularx}{0.20\textwidth}{c c c} 
		 	\toprule[0.2em]
				Method & PQ & PartPQ \\
				\midrule[0.15em]
			\rowcolor{gray!15}	 w PE &  61.6  & 57.4 \\
				w/o PE & 59.0  & 55.1  \\
				\bottomrule[0.1em]
	    \end{tabularx}
    } \hfill
    \subfloat[Effect on adding part annotations (anno).]{
        \label{tab:part_anno}
		\begin{tabularx}{0.16\textwidth}{c c} 
		\toprule[0.2em]
				Method &  PQ    \\  
				\midrule[0.15em]
			w/o part anno  &  61.2 \\
			\rowcolor{gray!15} w part anno & 61.6  \\
			    \bottomrule[0.1em]
		\end{tabularx}
    } \hfill
    \subfloat[Effect of adding boundary supervision, boundary loss (b-loss)]{
        \label{tab:boundary_loss}
		\begin{tabularx}{0.20\textwidth}{c c c} 
		\toprule[0.2em]
				Method &  PQ  & PartPQ  \\  
				\midrule[0.15em]
				\rowcolor{gray!15} baseline &  61.6  & 57.4 \\
			    + b-loss & 61.5  & 57.2  \\
			\bottomrule[0.1em]
		\end{tabularx}
    } \hfill
\end{table*}

%HERE
\noindent
\textbf{GFlops and Parameter Comparison on the CPP dataset.} 
The proposed unified model, Panoptic-PartFormer, is compact and efficient in terms of model parameters and run-time. 
Since the source code of~\cite{liang2020polytransform} is not publicly available, we estimate the lower bound by its baseline model~\cite{maskrcnn}.
As shown in Tab.~\ref{tab:gflops_parameter}, our model obtains a decrease of 55\%-60\% in GFlops and 65\%-70\% in parameters.
Compared with Panoptic-PartFormer, the proposed Panoptic-PartFormer++ only introduces an increase of 5\% GFlops and parameter cost. 
As shown in Tab.~\ref{tab:gflops_parameter}, our methods achieve significant improvements over previous models.

\noindent
\textbf{Results on the Pascal Panoptic Part Dataset.} We evaluate our method with the works on the Pascal Panoptic Part dataset. 
Tab.~\ref{tab:experiments_res_ppp} shows that, on different settings,
our methods achieve state-of-the-art results on both PQ and PartPQ with significant gains. 
For the ResNet backbone, our methods achieve \textbf{6\%-7\% gains} on PartPQ. 
Moreover, our ResNet101 model performs favorably against the prior work using larger backbones~\cite{qiao2021detectors}.
%
% Using Swin Transformer-base as backbone~\cite{liu2021swin}, our method obtains \textbf{47.4\% PartPQ}, which shows the generalization ability on a large model. 
% 
For our new proposed Panoptic-PartFormer++, under the same backbone, we can achieve \textbf{1.9\%-3.3\%} gains on PQ, \textbf{1.8\%} gains on PartPQ, and \textbf{3.0\%}  gains on PWQ.
For all metrics, this indicates the newly proposed models are good at handling all level segmentation predictions.

\noindent
\textbf{Scaling Up Panoptic-PartFormer++.} We analyze the performance
of Panoptic-PartFormer++ 
for both CPP and PPP datasets using the recent state-of-the-art backbones.
All the models are pre-trained on the COCO dataset and then fine-tuned on the CPP and PPP datasets. 
Different from previous works~\cite{degeus2021panopticparts,li2022panopticpartformer}, we do not use Mapillary data for pertaining. 
As shown in Tab.~\ref{tab:exp_scale_up}, we first explore various backbones on the CPP datasets, where we find that Convnext models~\cite{liu2022convnet} achieve the best results. 
We further explore the recent self-supervised pre-trained backbone BEIT-V2~\cite{bao2021beit}. 
However, we do not find extra gains in three metrics. 
The results show that a stronger pre-trained backbone is not critical for a better PPS model.
Then, we analyze the performance of our model on the PPP dataset
using various backbones.
%
%We find that different backbones perform differently for each metric. 
%
The model with Swin transformer performs better in terms of PartPQ and lower performance for PWQ. 
However, this is not the case with the Convnext models. 
These results show that the Convnext balances well for learning both part and scene segmentation tasks.  
Using the Convnext-base backbone, Panoptic-PartFormer++ model achieves state-of-the-art results.

%{\color{red} claim/conclusion not clear. }

\noindent
\textbf{Results on Cityscapes Panoptic Segmentation.}
We evaluate our method against the recent works on the Cityscapes Panoptic validation set. 
As shown in Tab.~\ref{tab:experiments_res_cityscapes}, our Panoptic-PartFormer also achieves favorable results against the previous works~\cite{li2020panopticFCN,cheng2020panoptic,zhang2021knet}. 
Panoptic-PartFormer++ achieves better results than the recent Mask2Former using the same backbone under the same settings. 
These results demonstrate the generalization ability of the proposed models on both panoptic segmentation and PPS tasks.

\begin{table}
\centering
\caption{\textbf{Experiment results on the Cityscapes Panoptic validation set.} $^*$ indicates using DCN~\cite{deformablev2}. All the methods use single-scale inference. Both K-Net and Mask2Former are pre-trained on COCO datasets. Thus, the performance is better than the original papers.}
\label{tab:experiments_res_cityscapes}
\begin{adjustbox}{width=0.45\textwidth}
\begin{tabular}{c c c c c}
\toprule[0.15em]
\textbf{Method}  &  Backbone  & $PQ$ & $PQ_{th}$ & $PQ_{st}$  \\ 
\midrule[0.15em]
UPSNet~\cite{xiong2019upsnet} & ResNet50 & 59.3 & 54.6 & 62.7 \\
SOGNet~\cite{yang2019sognet} & ResNet50 & 60.0 & {56.7} & 62.5 \\
Seamless~\cite{porzi2019seamless} & ResNet50 & 60.2 & 55.6 & 63.6 \\
Unifying~\cite{li2020unifying} & ResNet50 & 61.4 & 54.7 & 66.3 \\
Panoptic-DeepLab~\cite{cheng2020panoptic} & ResNet50 & 59.7 & - & - \\
Panoptic FCN$^*$~\cite{li2020panopticFCN} & ResNet50 & 61.4 & 54.8 & 66.6 \\
Panoptic FCN++~\cite{li2021fully} & Swin-large & 64.1 & 55.6 & 70.2 \\
K-Net~\cite{zhang2021knet} & ResNet50 & 61.2 & 52.4 & 66.8 \\
Mask2Former~\cite{cheng2021mask2former} & ResNet50 & 63.0 & 54.3  & 67.2 \\
Mask2Former~\cite{cheng2021mask2former} & Convnext-base & 67.5 &  61.1 & 71.8 \\
\hline
Panoptic-PartFormer & ResNet50 & 61.6 & 54.9 & 66.8 \\
Panoptic-PartFormer++ & ResNet50 &  63.6 & 57.5 & 68.1 \\
Panoptic-PartFormer & Swin-base & 66.6 & 61.7 & 70.3 \\
Panoptic-PartFormer++ & Swin-base & 68.0 & 61.9 & 72.0 \\
Panoptic-PartFormer++ & Convnext-base & 68.2 & 62.3 & 72.5 \\
\bottomrule
\end{tabular}
\end{adjustbox}

\end{table}

\begin{table*}[!t]
     \centering
    \caption{\textbf{Ablation studies and analysis on Panoptic-PartFormer++ using the CPP validation set.} GMC: Global Masked Cross attention. PMC: Part Masked Cross attention. DD: Decoupled Decoder. DPH: Dense Prediction Head from the auxiliary head. ResNet50 is adopted as the backbone. 
    }
    \label{tab:ablation_ppformer_plus}
    \subfloat[Effect of Each Component]{
        \small
        \label{tab:effect_component_ppformer_plus}
	    \begin{tabularx}{0.42\textwidth}{c c c c c c c c} 
		        				\toprule[0.15em]
    	DD & GMC & PMC & PartPQ & PWQ & SSQ & PSQ \\
            \midrule[0.15em]
    	\rowcolor{gray!15} \checkmark &  \checkmark & \checkmark &  59.2 & 62.1 &  49.6 &  27.5  \\	
        - &  \checkmark & \checkmark & 58.0 & 60.2 & 49.5 & 25.3  \\	
        \checkmark &  \checkmark & - & 58.5 & 60.5 & 49.3 & 26.0 \\
           \bottomrule[0.1em]
	    \end{tabularx}
    } \hfill
    \subfloat[Query Reasoning Design]{
        \label{tab:query_reasoning_ppformer_plus}
		\begin{tabularx}{0.24\textwidth}{c c c} 
			\toprule[0.15em]
		  Setting & PartPQ & PWQ \\
			\midrule[0.15em]
		 Joint Reasoning & 58.2 & 59.9 \\
          Part First & 58.3 & 60.5 \\
         \rowcolor{gray!15} Global First & 59.2 & 62.1 \\
          \bottomrule[0.1em]
		\end{tabularx}
    } \hfill
    \subfloat[Choice Of PPS Fusion.]{
        \label{tab:effect_dph_ppformer_plus}
		\begin{tabularx}{0.29\textwidth}{c  c c} 
		  \toprule[0.20em]
		Merge Method & PartPQ & PWQ \\
		  \midrule[0.15em]
		\rowcolor{gray!15}  with Part Query & 59.2 & 62.1 \\
             with DPH &  57.4 & 59.2 \\ 
            Merged DPH/Part Query &  58.4 & 61.5 \\    
		\bottomrule[0.1em]
		\end{tabularx}
    } \hfill
\end{table*}

%HERE
\subsection{Ablation Study and Analysis}
\label{sec:ablation}
We present ablation studies and design analysis of the Panoptic-PartFormer and Panoptic-PartFormer++ models. 
We first present the core design and analysis of Panoptic-PartFormer (Sec.~\ref{sec:ablation_pppformer}). 
Note that since both models share the same meta-architecture design, several designs are the same. 
Thus, we only explore the core design of Panoptic-PartFormer++ in Sec.~\ref{sec:ablation_pppformer_plus}. 
We then present the comparisons with Mask-Transformer-like baselines.
%(shown in Fig.~\ref{fig:meta_architecture}(a)).

\subsubsection{Ablation on Panoptic-PartFormer}
\label{sec:ablation_pppformer}
\noindent
\textbf{effectiveness of each component.} As shown in Tab.~\ref{tab:effect_component}, we show the effectiveness of each component of our framework by removing it from the original design. 
Removing the Decoupled Decoder (DD) results in a 1.4\% drop in PartPQ. 
Removing Dynamic Convolution (DC) or Self Attention (SA) leads to a large drop in PQ, which means both are important for the interaction between queries and corresponding query features. 
Decreasing the stage number to 1 also leads to a significant drop. Performing more interaction results in a more accurate feature location for each query, which is the same as previous works~\cite{zhang2021knet,peize2020sparse}.

\noindent
\textbf{Part query depends more on the thing query?} With our framework, we can easily analyze the relationship between the stuff query, thing query, and part query. 
We present several ways of reasoning and fusing different queries. 
From intuitive thought, part information is more related to thing query. 
We design two different query interaction methods, shown in Tab.~\ref{tab:query_reasoning}. 
For separate reasoning, we use DD and SA on two query pairs: the stuff-thing query and the thing-part query. 
For sequential reasoning, we perform DD and SA with the thing-part query first and the stuff-thing query second. 
However, we find the best model is the joint reasoning, which is the default setting described in Sec.~\ref{sec:ppformer_eccv}. 
These results show that better part segmentation needs the whole scene context rather than thing features only.

\noindent
\textbf{Joint query modeling or separate modeling on part queries?} Using the same settings as PanopticFPN~\cite{kirillov2019panopticfpn}, we use the semantic-FPN-like model for part segmentation. 
Dense Prediction (DP) is the baseline method shown in Fig.~\ref{fig:teaser_01}(b). 
We adopt the same merging process for panoptic segmentation and part segmentation. 
As shown in Tab.~\ref{tab:dp_vs_jq}, our joint query-based method achieves better results and outperforms the previous dense prediction-based approach and an improved model~\cite{deeplabv3}. 
These results indicate that joint learning greatly benefits part segmentation and demonstrate the effectiveness of the proposed framework.  

%HERE
\noindent
\textbf{Aligned decoder is more important for part segmentation.} As shown in Tab.~\ref{tab:aligned_decoder}, using the aligned part decoder results in better PartPQ, especially for the things with parts. Adding both paths with the aligned decoder does not bring extra gain. 
These results show that part segmentation needs more detailed information, while thing and stuff predictions do not rely on it. 
%
%Thus, we also keep this design for the Panoptic-PartFormer++.

\noindent
\textbf{Necessity of Positional Encoding on $X_{p}$ and $X_{u}$.} 
The results in Tab.~\ref{tab:pos_enc} show that removing positional encoding leads to inferior results on both PQ and PartPQ, indicating the importance of position information~\cite{wang2020solov2,zhang2021knet,detr}.
%
%Thus, we keep this design for both Panoptic-PartFormer and Panoptic-PartFormer++.

\noindent
\textbf{Will boundary supervision help for part segmentation?} In Tab.~\ref{tab:boundary_loss}, we also add boundary supervision for part segmentation, where we use the dice loss~\cite{milletari2016v} and binary cross entropy loss. 
Although boundary supervision may seem to be useful for part segmentation~\cite{li2020improving,kirillov2019pointrend}, 
there is no performance gain observed.
As our mask is generated from the aligned decoder, it already contains detailed information. 
These results motivate us to focus on the relationship between the scene, and part features rather than modeling part segmentation solely.

\noindent
\textbf{Will joint training help for panoptic segmentation?} As shown in Tab.~\ref{tab:part_anno}, joint learning benefits the panoptic segmentation baseline. 
However, the performance gains are limited since both thing, and stuff predictions do not need much detailed information.

%HERE
\subsubsection{Ablation on Panoptic-PartFormer++}
\label{sec:ablation_pppformer_plus}
% We evaluate the Whole Seg Quality / Part Seg Quality. 
% Ablation on Each Component
% Ablation on Global-Part Masked Cross Attention Design.
% Comparison on Different Backbone. 
\noindent
\textbf{Effectiveness of Each Component.} In Tab.~\ref{tab:effect_component_ppformer_plus}, we demonstrate the effectiveness of each component by removing it from the proposed model. 
Removing the decoupled decoder leads to a 1.2\% PartPQ drop and a 1.8\% PSQ drop from the Panoptic-PartFormer++ model. 
These results indicate the decoupled design is a key component for both Panoptic-PartFormer and Panoptic-PartFormer++. 
Removing the part masked cross attention module leads to a 1.5\% PSQ drop. 
This shows the effectiveness of the extra part query attention with part features. 
In both cases, the scene segmentation results (SSQ) are the same. 
The results indicate that part attention and decoupled decoder design do not affect panoptic segmentation.
Moreover, these results also indicate the interpretability of our proposed PWQ metric, where we achieve improvements mainly from the better part segmentation quality.

\noindent
\textbf{Ablation on Global-Part Masked Cross Attention Design.} In Tab.~\ref{tab:query_reasoning_ppformer_plus}, we analyze the query interaction design in Panoptic-PartFormer++. 
We use three different attention methods, including joint reasoning, part-first reasoning, and global-first reasoning. 
For joint reasoning, we perform masked cross-attention on three queries with the corresponding scene features. 
For part-first attention module, we first perform part cross attention for part query and part feature, and then we perform the joint reasoning on three queries. 
In the global-first attention module, we first perform joint reasoning, and then we perform the part cross attention. 
Different from the results in Tab.~\ref{tab:query_reasoning}, after adopting the masked cross attention, we find adopting global-first leads to the best results. 
This is because masked attention contains more fine-grained information, and it can be improved with the guidance of joint global scene query learning. 
In the appendix, we present more detailed ablation studies in our global part-masked cross-attention design.
Specifically, we present three more different designs on the location of the refined part query, where our proposed global reasoning and local refined design lead to better results. 
%
%Due to the space limitation of the main paper, we refer the reader to look at our appendix.

\noindent
\textbf{Ablation on Extra Part Dense Prediction.} Our Panoptic-PartFormer++ also adopts an extra semantic part segmentation head during training for part features.
Removing such supervision leads to about 0.3\% PartPQ drop and 0.8\% PWQ drop.
%We do not list it in a table to save space. 
This result means the decoupled design can be further improved by adding task-specific loss.

\noindent
\textbf{Ablation on Final PPS merging.} Since our Panoptic-PartFormer++ has an extra dense semantic segmentation prediction head for part feature learning, we also explore the results of its part segmentation quality.
As shown in Tab.~\ref{tab:effect_dph_ppformer_plus}, replacing the dense prediction results leads to a significant drop. 
This indicates the effectiveness of our unified modeling for PPS. 

\begin{figure}[!t]
	\centering
	\includegraphics[width=1.0\linewidth]{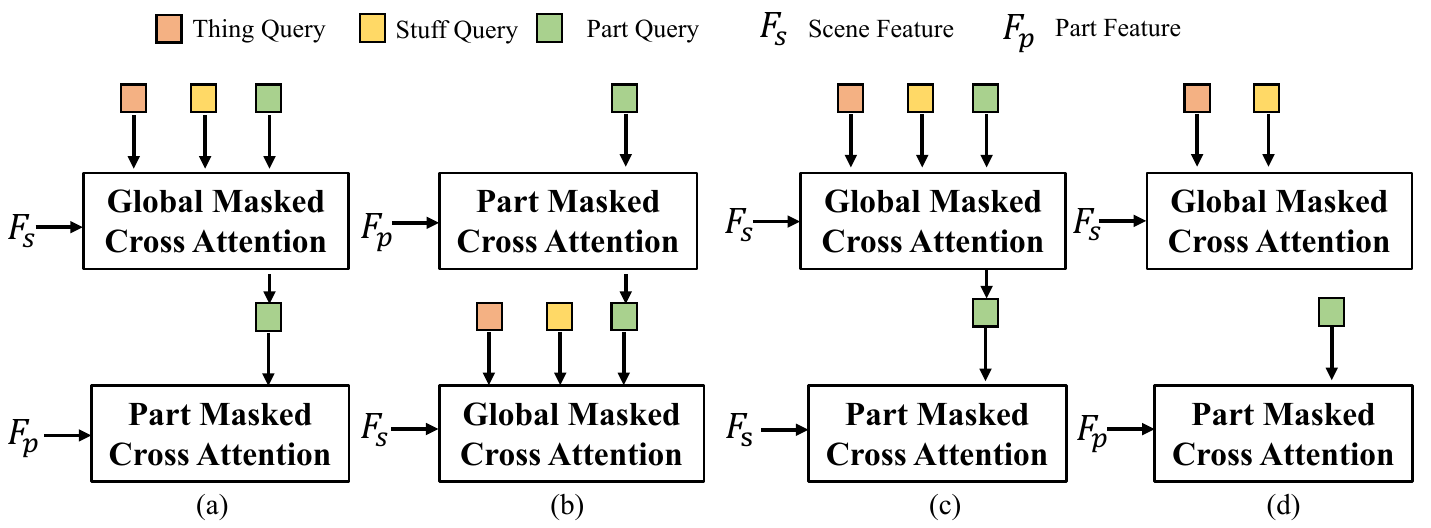}
	\caption{Ablation Study on Global-Part Masked Cross Attention Design. (a), Our Proposed Method. (b), Local Part Query Cross Attention First. (c), Scene Feature for Part Query Cross Attention. (d), Solo Local Part Query Cross Attention.}
	\label{fig:ablation_head_design}
\end{figure}

\begin{table}[!t]\setlength{\tabcolsep}{6pt}
   \caption{\textbf{Comparison of Global-Part Masked Cross Attention Design in Fig.~\ref{fig:ablation_head_design}} All models use ResNet50 as backbone and meta-architecture is PanopticPartFormer++.}
    \label{tab:more_ablation_on_global_part_desgin}
   \centering
   \begin{adjustbox}{width=0.40\textwidth}
	    \begin{tabular}{l c c c} 
		    \toprule[0.15em]
    		 Method & PQ & PartPQ & PWQ    \\
    		\midrule[0.15em]
         \rowcolor{gray!15}  Global First (Fig.~\ref{fig:ablation_head_design} (a)) & 63.6 & 59.2 & 62.1\\
          Part First (Fig.~\ref{fig:ablation_head_design} (b)) & 63.3 & 58.3 & 60.5 \\
         Using Scene Features (Fig.~\ref{fig:ablation_head_design} (c)) & 63.6 & 58.0 & 61.5 \\
          Solo Local Part Query  (Fig.~\ref{fig:ablation_head_design} (d)) & 62.8 & 58.3 & 60.0 \\
          \bottomrule[0.1em]
	\end{tabular}
    \end{adjustbox}
\end{table}

\begin{table}[!t]\setlength{\tabcolsep}{6pt}
   \caption{\textbf{Exploration Multi-Task Dense Prediction Approaches for PPS.} All models use ResNet50 as backbone and meta-architecture is PanopticPartFormer++.}
    \label{tab:ablation_approach_in_mdp}
   \centering
   \begin{adjustbox}{width=0.40\textwidth}
	    \begin{tabular}{l c c c c} 
		    \toprule[0.15em]
    		 Method & PQ & PartPQ & PWQ  & GFlops \\
    		\midrule[0.15em]
         \rowcolor{gray!15}  Our Default Baseline & 63.6 & 59.2 & 62.1 & 210.5 \\
        + MQFormer~\cite{xu2023mqformer} & 62.3 & 56.4 & 60.2 & 223.4 \\
        + TaskPrompter~\cite{taskprompter2023} & 63.0 & 58.2 & 61.2 & 215.6 \\
                 \bottomrule[0.1em]
	\end{tabular}
    \end{adjustbox}
\end{table}

\begin{table}[!t]\setlength{\tabcolsep}{6pt}
   \caption{\textbf{Compared with Single Task Baseline.} All models use ResNet50 as backbone. mIoU evaluates part segmentation.}
    \label{tab:ablation_single_baseline}
   \centering
   \begin{adjustbox}{width=0.40\textwidth}
	    \begin{tabular}{l c c c c} 
		    \toprule[0.15em]
    		 Method & PQ & PartPQ & PWQ  & mIoU \\
    		\midrule[0.15em]
         \rowcolor{gray!15}  Our Default Baseline & 63.6 & 59.2 & 62.1 & 52.0 \\
         Single Part Segmentation & -  &  - &  - & 50.2 \\
         Single Panoptic Segmentation &  63.2 & - & - & - \\
         \bottomrule[0.1em]
	\end{tabular}
    \end{adjustbox}
\end{table}

\begin{table}[!t]\setlength{\tabcolsep}{6pt}
   \caption{\textbf{Comparison of Meta-Architecture Design in Fig.~\ref{fig:meta_architecture} (a) and (b) on the CPP validation set.} Both models achieve better results on all three metrics.}
    \label{tab:more_exp_with_meta_maskformer}
   \centering
   \begin{adjustbox}{width=0.40\textwidth}
	    \begin{tabular}{l c c c c} 
		    \toprule[0.15em]
    		 Method & PQ & PartPQ & PWQ  & GFlops   \\
    		\midrule[0.15em]
            K-Net + Part Query &  60.8 & 56.0 &  58.1  &  183.2 \\
                      \rowcolor{gray!15}  Panoptic-PartFormer &   61.6  &  57.5  &  60.1  &  185.8  \\
                    \hline
            Mask2Former + Part Query & 62.7 & 58.2 & 60.2 & 210.5\\
           \rowcolor{gray!15}  Panoptic-PartFormer++ & 63.4   &  59.2  &  62.1 &  215.4 \\
          \bottomrule[0.1em]
	\end{tabular}
    \end{adjustbox}
\end{table}

\begin{table}[!t]\setlength{\tabcolsep}{6pt}
	\centering
 	\caption{\textbf{Effect Of COCO pretraining.} We use ResNet50 as the backbone on the CPP validation set. Different from Tab.~\ref{tab:ablation_ppformer} and Tab.~\ref{tab:ablation_ppformer_plus}, all the methods use the same training setting. }
  	\label{tab:coco_pretrain}
	\begin{adjustbox}{width=0.50\textwidth}
	\begin{tabular}{l c c c c}
				\toprule[0.15em]
	Method & COCO pre-train & PQ & PartPQ  & PWQ \\
				\toprule[0.15em]
	\textbf{Panoptic-PartFormer} &  - & 54.5 & 57.8 &   54.6   \\
        \textbf{Panoptic-PartFormer} & \checkmark &  57.4 &  61.6 & 60.5  \\
	\textbf{Panoptic-PartFormer++} & - & 61.4 &  57.5 & 58.8 \\
        \textbf{Panoptic-PartFormer++} & \checkmark & {63.6} & {59.2} & 62.1 \\ 
	\bottomrule[0.1em]
	\end{tabular}
	\end{adjustbox}
\end{table}

\noindent
\textbf{Ablation on Global-Part Masked Cross Attention Design.} We show detailed ablation designs of our Panoptic-PartFormer++ head in Fig.~\ref{fig:ablation_head_design}. Fig.~\ref{fig:ablation_head_design}(a) is our default design.
Fig.~\ref{fig:ablation_head_design}(b) is Part Query Cross attention First design. 
Fig.~\ref{fig:ablation_head_design}(c) uses scene features $F_{s}$ for both global and local cross-attention. 
Fig.~\ref{fig:ablation_head_design}(d) is solo Part Query Cross Attention without Global Refinement. We report the main results in Tab.~\ref{tab:more_ablation_on_global_part_desgin}. 
The proposed method achieves the best result on all three metrics. Using part query first leads to inferior results on Part Segmentation. 
This is because, after global cross-attention with the scene feature, the local part information is lost. Using scene features also leads to bad results, which demonstrates the effectiveness of our decoupled decoder head design. 
Part segmentation needs more fine-grained feature representation. 
Moreover, using solo local part queries without global reasoning has worse results. 
This indicates the part segmentation still needs the global prior, which shares a similar finding in Tab.~\ref{tab:ablation_ppformer} (c).

\noindent
\textbf{Explore Methods in Multi-Task Dense Prediction For PPS.} 
In Tab.~\ref{tab:ablation_approach_in_mdp}, we also explore two representative approaches~\cite{xu2023mqformer,taskprompter2023} for the cross-task feature learning into our PanopticPartFormer++ framework. In particular, we use their open-source implementation~\footnote{https://github.com/prismformore/Multi-Task-Transformer, https://github.com/yangyangxu0/MQTransformer }. We merge their operation into the decoupled feature decoder but keep the remaining unchanged. From the results of Tab.~\ref{tab:ablation_approach_in_mdp}, though increasing the GFLops, we do not find any performance gains over our baselines. This is because feature-level interaction is not suitable for PPS tasks, while the entity-level interaction between the scene query feature and part query feature is more important.

\noindent
\textbf{Comparison with Single Head Baseline.} We also verify the effectiveness of multi-task co-training in our framework in Tab.~\ref{tab:ablation_single_baseline}. In particular, we remove the scene head or part head and the remaining components are the same. 
From that table, we find significant improvement in part segmentation and nearly no improvements in panoptic segmentation, which shares the same conclusion in Tab.~\ref{tab:ablation_ppformer} (f) of PanopticPartFormer. Joint co-training leads to better part segmentation results.

\subsubsection{Comparison with Baseline Methods}
\label{sec:meta_architecutre_exp}
% K-Net mask-former with our Meta-Architecture 
% Mask2Former with our Meta-Architecutre.
% Effect of COCO-pretraining for CPP and PPP.

\noindent
\textbf{Compared with MaskFormer Meta-Architecture}  Tab.~\ref{tab:more_exp_with_meta_maskformer} shows the evaluation results of the proposed and MaskFormer models (Fig.~\ref{fig:meta_architecture}(a)). 
All the methods use ResNet-50 as the backbone. Although both works~\cite{zhang2021knet,cheng2021mask2former} achieve better results than the previous separate baselines~\cite{degeus2021panopticparts}. 
Our methods achieve better results with fewer extra GFlops costs. 
These results demonstrate the effectiveness of our Meta-Architecture (Fig.~\ref{fig:meta_architecture}(b)). 
%
%In addition,  the proposed architecture is \textit{not a simple extension} of MaskFormer with part query. 
%
In addition, our methods provide a unified and decoupled view for the PPS task, which is significantly different from the MaskFormer study. 

\noindent
\textbf{Effect of COCO-pretraining.} In Tab.~\ref{tab:coco_pretrain}, we demonstrate the effect of COCO pertaining on the CPP dataset. 
The performance of both Panoptic-PartFormer and Panoptic-PartFormer++ drops without COCO-pretraining.
However, the new proposed Panoptic-PartFormer++ is less sensitive to COCO-pretraining. 
These results indicate that Panoptic-PartFormer++ is also a data-efficient model.
For the effect of COCO-pretraining on the PPP dataset, we refer the reader to the appendix.

\begin{figure}[t!]
	\centering
	\includegraphics[width=1.0\linewidth]{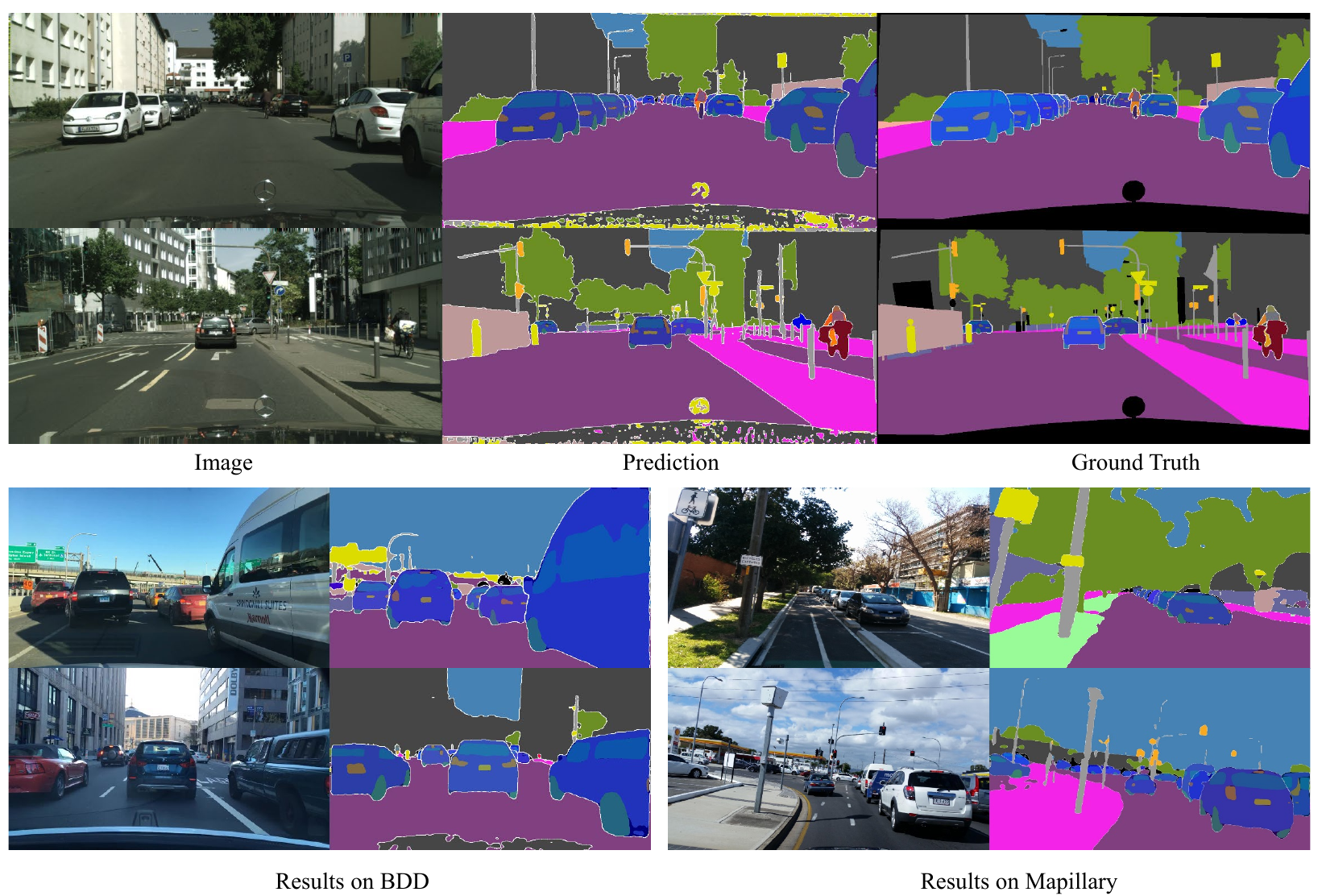}
	\caption{Visualization of our Panoptic-PartFormer. Top: results on Cityscapes PPS validation set. Bottom left: prediction on BDD dataset~\cite{yu2020bdd100k}. Bottom right: prediction on Mapillary dataset~\cite{neuhold2017mapillary}. Best view it on screen.}
	\label{fig:exp_vis}
\end{figure}

\subsection{Visualization Analysis}
\label{sec:vis_analysi}
% Visualization of PPFormer 
% Visual Improvements of PPFormer++.
\begin{figure}[!t]
	\centering
	\includegraphics[width=1.0\linewidth]{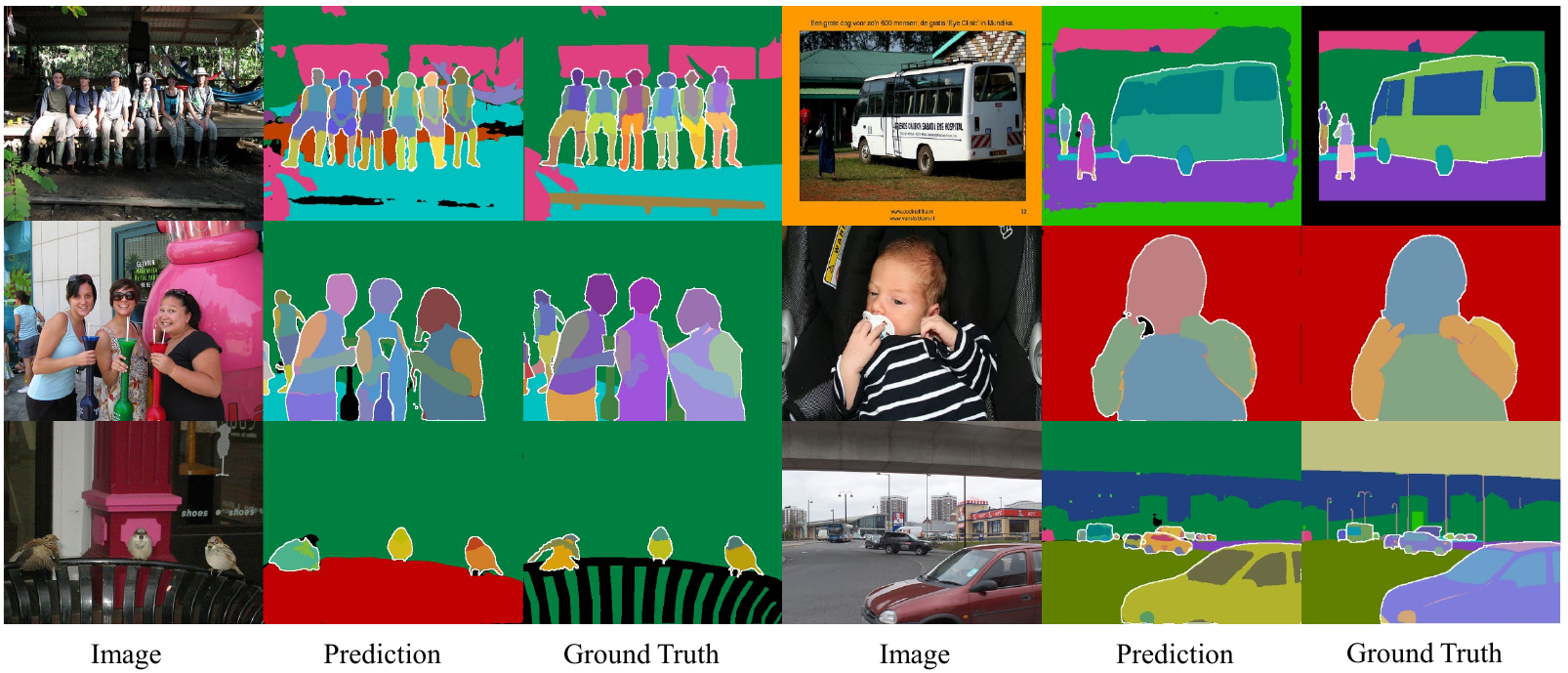}
	\caption{Visualization results on the Pascal Context Panoptic Part validation set using Panoptic-PartFormer. Note that stuff classes have the same color, while thing and part classes are not. Best view it on screen.}
	\label{fig:vis_results_pascal_pss}
\end{figure}

\noindent
\textbf{Visualization and Generalization With Panoptic-PartFormer.} We present several visualization examples using our model on Cityscapes PPS validation set. 
Moreover, we also visualize several examples on the Mapillary dataset~\cite{neuhold2017mapillary} and BDD dataset~\cite{yu2020bdd100k} to show the generalization ability of our method.
As shown in the first row of Fig.~\ref{fig:exp_vis}, our method achieves considerable results. 
Moreover, on the Mapillary~\cite{neuhold2017mapillary} and BDD datasets~\cite{yu2020bdd100k}, which do not have part annotations, our method can still work well as shown in the last row of Fig.~\ref{fig:exp_vis}. 
In addition, we also visualize the results on the PPP datasets in Fig.~\ref{fig:vis_results_pascal_pss}. 
The first two rows show the crowded human scene and outdoor scene. 
Both cases show that our model can obtain convincing results. 
The last row shows the small object cases. The failure cases are due to tiny objects, including their inner parts. 

\begin{figure}[!t]
     \centering
     \includegraphics[width=0.75\linewidth]{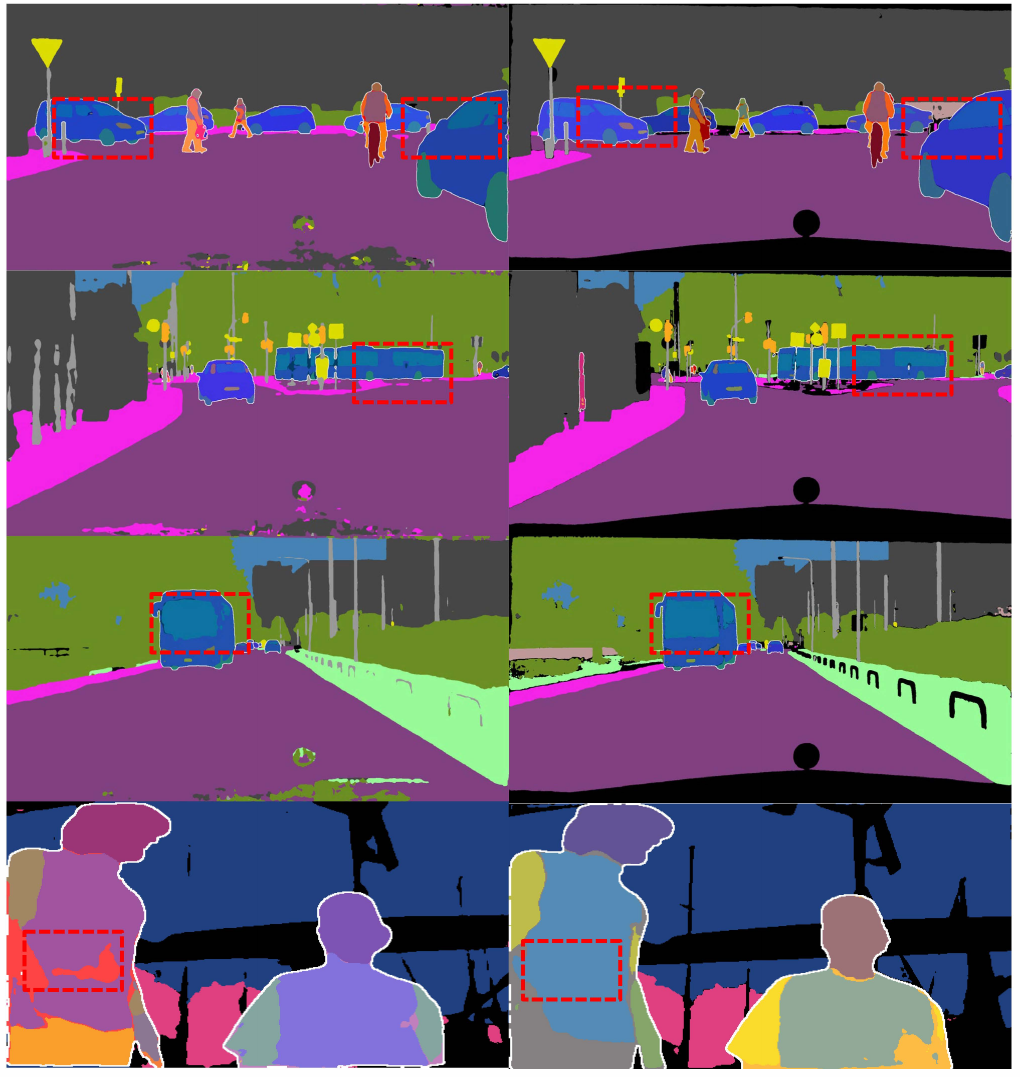}
    \caption{Visualization improvements on the CPP and PPP datasets. Left: Panoptic-PartFormer. Right: Panoptic-PartFormer++. Both models use the ResNet-50 backbone. Shown in the red boxes, Panoptic-PartFormer++ achieves better part segmentation results on both CPP and PPP datasets.}
    \label{fig:vis_improvements}
\end{figure}

\noindent
\textbf{Visual Improvements by Panoptic-PartFormer++.} We show some sample results of PanopticPartFormer++ in Fig.~\ref{fig:vis_improvements}. 
Overall, Panoptic-PartFormer++ has better part segmentation quality (better boundaries, aligned semantic consistency within the part) on both the CPP and PPP datasets, as shown in the red boxes of Fig.~\ref{fig:vis_improvements}. 
We present more visualization examples in the appendix.

\begin{figure}[!t]
     \centering
     \includegraphics[width=0.85\linewidth]{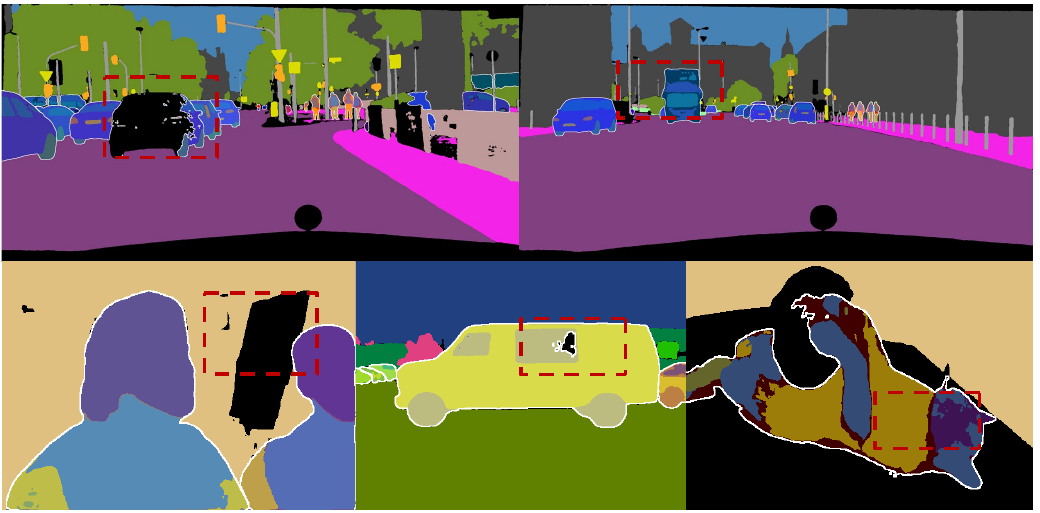}
    \caption{Failure cases on the CPP (top) and PPP (bottom) datasets.}
    \label{fig:fail_cases}
    \vspace{-2mm}
\end{figure}

\noindent
\textbf{Failure Cases Analysis.} In Fig.~\ref{fig:fail_cases}, we show some failure cases of the Panoptic-PartFormer++ models. 
As shown in the red boxes, there are two issues: wrong background prediction during PPS merging and inconsistency prediction within the part. 
These results suggest that a better fusion procedure or a new part segmentation branch may be the future research direction of PPS. 
We present more failure cases in the appendix.

	\section{Conclusion}
This paper addresses the panoptic part segmentation task by designing a unified and decoupled model named Panoptic-PartFormer. 
We present a decoupled meta-architecture and perform joint reasoning along with thing, stuff, and part queries and their corresponding features. 
Using this model, we find the unbalanced issues of the PartPQ metric. 
To handle this problem, we present a balanced and error-decoupled metric from two perspectives: pixel-region and part-scene. 
Then, we propose a new enhanced model based on our meta-architecture and Mask2Former, named Panoptic-PartFormer++. 
We design global-part masked cross attention to perform interactions between scene features and part features effectively. 
We further conduct extensive experiments on two widely used benchmark datasets.
Furthermore, our models achieve state-of-the-art results on both datasets with much fewer GFlops and parameters than existing approaches.
	
% \section{}
\appendix
\section{Appendix}

\subsection{More Detained Experiment Settings.}

\noindent
\textbf{For Mapillary~\cite{neuhold2017mapillary} dataset pretraining}, we adopted the settings used in Panoptic-Deeplab~\cite{cheng2020panoptic}, including multiscale training with scales ranging from 1.0 to 2.0 of the original image, random crops of 1024 $\times$ 2048 patches, and horizontal flips. Due to limited computational resources, we trained for 240 epochs, although we believe more training iterations could improve results. We used the Mapillary dataset for pretraining to allow for a fair comparison with previous work~\cite{degeus2021panopticparts,ocrnet,mohan2021efficientps}, which all utilized Mapillary pretraining for improved results. Notably, we only used the Mapillary dataset for pretraining our Panoptic-PartFormer model, and we \textbf{do not} pre-train Panoptic-PartFormer++ on the Mapillary dataset.

\noindent
\textbf{For COCO~\cite{coco_dataset} dataset pretraining}, we utilized multiscale training as done in prior work~\cite{detr}, where input images are resized so that the shortest side is within the range of 480 to 800 pixels, and the longest side is at most 1333 pixels. We also incorporated random crop augmentations during training by randomly cropping training images with a probability of 0.5 to a rectangular patch, which is then resized to dimensions of 800 pixels in height and 1333 pixels in width. All models were trained for a total of 36 epochs. Additionally, for Panoptic-PartFormer++, we employed the Mask2Former pretraining process, using a large-scale jitter setting and training the model for 50 epochs.

\begin{table}[!t]\setlength{\tabcolsep}{6pt}
   \caption{\textbf{More ablation of Global-Part Masked Cross Attention Design.} The last two rows use Panoptic-PartFormer++ architecture.}
    \label{tab:more_ablation_on_global_design}
   \centering
   \begin{adjustbox}{width=0.40\textwidth}
	    \begin{tabular}{l c c c} 
		    \toprule[0.15em]
    		 Method & PQ & PartPQ & PWQ    \\
    		\midrule[0.15em]
         Panoptic-PartFormer & 61.6 & 57.4 & 60.5 \\
         \rowcolor{gray!15} Masked Cross Attention & 63.6 & 59.2 & 62.1\\
         Mask Pooling based Dynamic Convolution~\cite{peize2020sparse,zhang2021knet} & 62.0 & 58.5 & 61.5 \\
         
          \bottomrule[0.1em]
	\end{tabular}
    \end{adjustbox}
\end{table}

\begin{table*}[!h]
\centering
\begin{adjustbox}{width=1.0\textwidth}

\begin{tabular}{ l | c c c c c c c c c c c c c c c c c c c | c}
	\toprule[0.15em]
		Method & road & swalk & build & wall & fence & pole & tlight & sign & veg. & terrain & sky & person & rider & car & truck & bus & train & mbike & bike & mean PartPQ \\
    \toprule[0.15em]
		Previous hybrid models & 98.3 & 80.4 & 90.3 & 37.7 & 44.0 & 63.4 & 58.5 & 74.5 & 90.9 & 41.1 & 88.8 & 44.1 & 45.3 & 53.3 & 36.4 & 49.7 & 67.9 & 50.2 & 51.6 & 61.4  \\
		Panoptic-PartFormer & 98.0 & 78.2 & 89.5 & 43.5 & 44.4 & 59.3 & 59.5 & 74.4 & 90.5 & 45.8 & 90.0 & 46.0 & 45.9 & 50.2 & 35.1 & 51.0 & 75.4 & 50.5 & 50.1 & 61.9 \\
	\bottomrule		
	\end{tabular}
\end{adjustbox}
\vspace{0pt}
\caption{Detailed experiment results on Cityscapes Panoptic validation set}
\label{tab:experiments_res_cityscapes_details}
\end{table*}

\begin{figure*}[!t]
	\centering
	\includegraphics[width=1.0\linewidth]{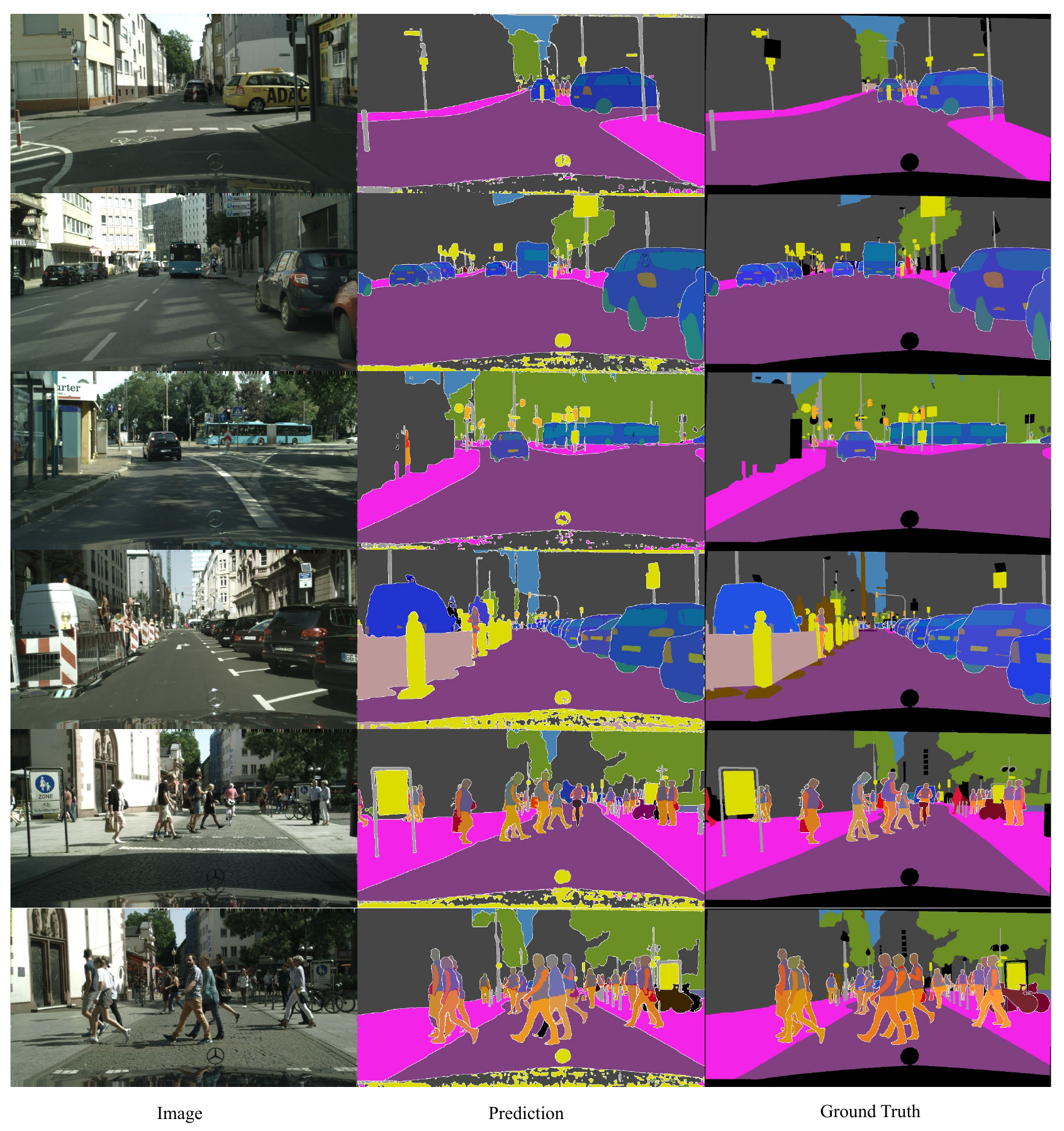}
	\caption{More visualization results on Cityscapes Panoptic Part validation set. Best viewed in color and by zooming in. Black regions are ignored during the evaluation. We show examples of driving cars and buses in the first four rows.}
	\label{fig:sub_vis_results_city_pss}
\end{figure*}

\subsection{More Experiment Results}

\noindent
\textbf{More Ablation on Global Mask Cross Attention.} In Tab.~\ref{tab:more_ablation_on_global_design}, we replaced the cross-attention mechanism for all three queries with mask-pooling-based dynamic convolution. As our model has a stronger backbone than the previous Panoptic-PartFormer, we observed improvements in all three metrics. However, when using masked cross-attention, we observed a drop in performance for all three metrics. This indicates that more fine-grained cross-attention results in better segmentation quality for both PS and PPS. Adopting pooling-based attention resulted in missing details in both scene features and part features.

\noindent
\textbf{Detailed Results on Cityscapes PPS.} In Tab.~\ref{tab:experiments_res_cityscapes_details}, we provide a detailed breakdown of PartPQ results for each class. Our method outperforms previous methods in various classes, such as train, wall, sky, person, rider, etc.

\noindent
\textbf{More Visualization results on Cityscapes PPS.} Fig.~\ref{fig:sub_vis_results_city_pss} displays additional examples of Cityscapes Panoptic Part Segmentation. The first four rows depict results on a road driving scene, while the last two rows feature a crowded human scene. Our Panoptic-Partformer model performs well in both scenarios, and we used the Swin-base model for visualization.

\begin{figure*}[!t]
	\centering
	\includegraphics[width=1.0\linewidth]{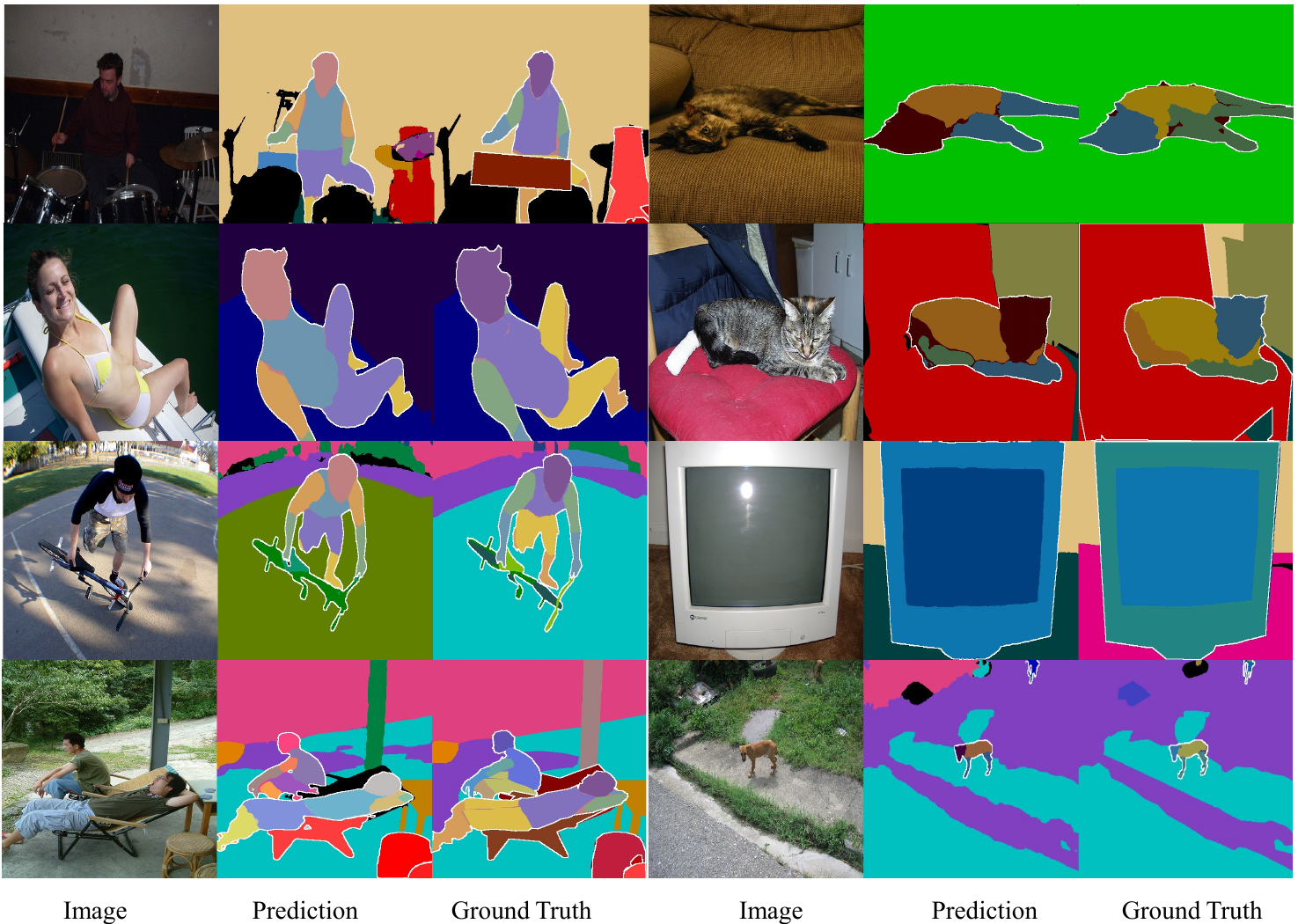}
	\caption{More visualization results on Pascal Context Panoptic Part validation set. Best viewed in color and by zooming in. \textit{Note that stuff classes have the same color, while thing and part classes are not.}}
	\label{fig:sub_vis_results_pascal_pss}
\end{figure*}

\noindent
\textbf{More Visualization results on Pascal Context PPS.} Fig.~\ref{fig:sub_vis_results_pascal_pss} showcases additional examples of the Pascal Context Panoptic Part Segmentation dataset. The figures on the left depict human scenes, including a crowded scenario, while the figures on the suitable feature scenes with non-human parts.

\begin{figure*}[!t]
	\centering
	\includegraphics[width=1.0\linewidth]{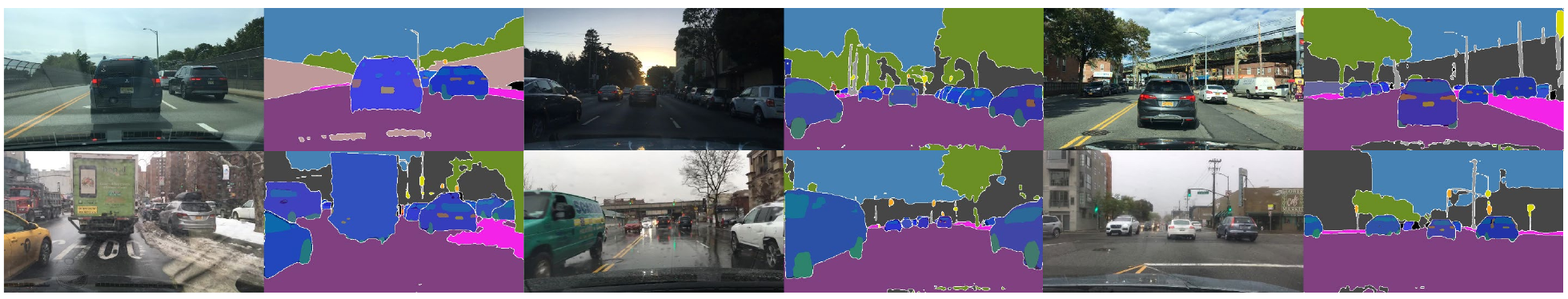}
	\caption{More visualization results on BDD dataset. Best viewed in color and by zooming in. We use the color map of Cityscapes for visualization. The first row shows the normal driving cases, while the second row shows the driving cases with different weather scenarios (such as rain).}
	\label{fig:sub_vis_results_bdd_pss}
\end{figure*}

\begin{figure*}[!t]
	\centering
	\includegraphics[width=1.0\linewidth]{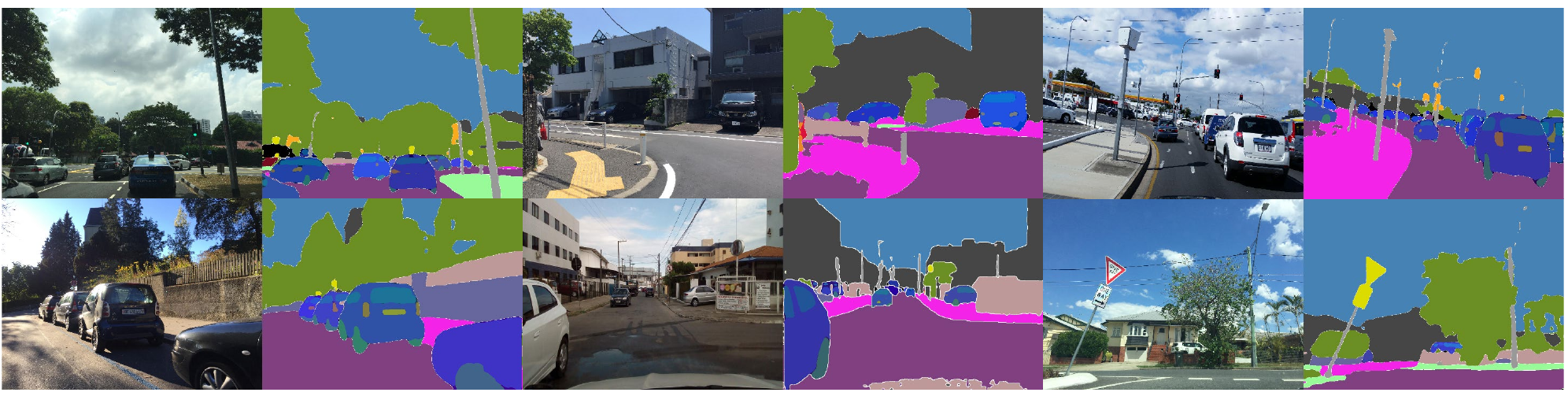}
	\caption{More visualization results on Mapillary dataset. Best viewed in color and by zooming in. We use the color map of Cityscapes for visualization. Both rows show the generalization ability of our approaches.}
	\label{fig:sub_vis_results_mapillary_pss}
\end{figure*}

\noindent
\textbf{More generalization results on Mapillary and BDD.} In Fig.~\ref{fig:sub_vis_results_bdd_pss}, we give more visual results on generalization on BDD datasets. The second row shows the scene with rainy weather. Our Panoptic-Partformer still works well, which shows both the robustness and generalization of our method. All the figures are obtained from our Cityscapes models \textit{without} training on the BDD dataset. In Fig.~\ref{fig:sub_vis_results_mapillary_pss}, we present more results on Mapillary datasets.

\begin{figure*}[!t]
	\centering
	\includegraphics[width=0.80\linewidth]{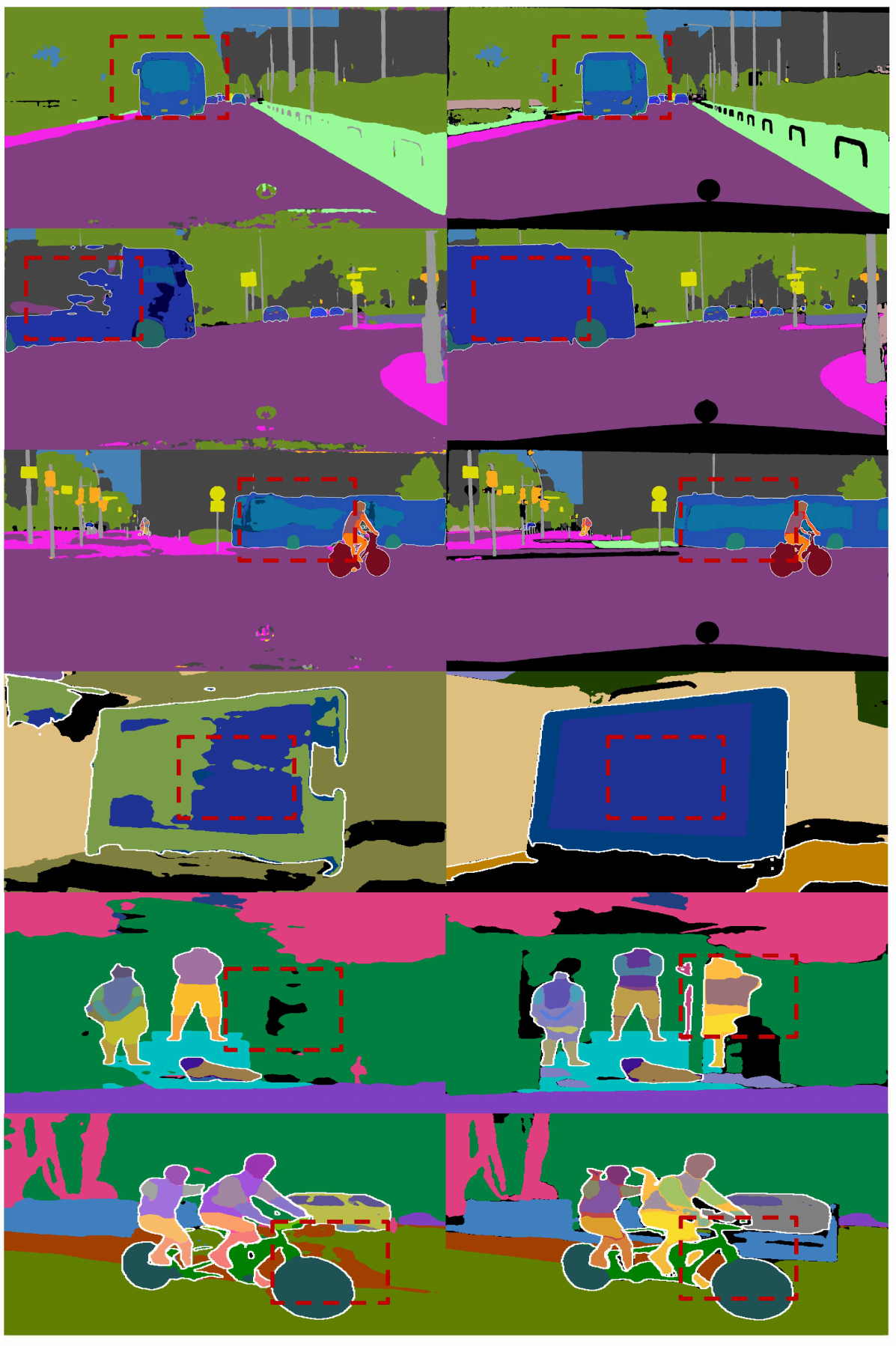}
	\caption{Visualization improvements on CPP and PPP datasets. Left: Panoptic-PartFormer. Right: Panoptic-PartFormer++. Both models use the ResNet-50 backbone. Shown in the red boxes, Panoptic-PartFormer++ achieves better part segmentation results on both CPP and PPP datasets.}
	\label{fig:more_visual_improvements}
\end{figure*}

\noindent
\textbf{More Visual Improvement results on Cityscapes PPS and Pascal Context PPS.} We show more results of PanopticPartFormer++ in Fig.~\ref{fig:more_visual_improvements}. Compared to Panoptic-PartFormer, our new proposed Panoptic-PartFormer++ has better part segmentation quality, including better boundaries, aligned semantic consistency within the part, and better background stuff segmentation on both CPP and PPP datasets.

\begin{figure*}[!t]
	\centering
	\includegraphics[width=0.80\linewidth]{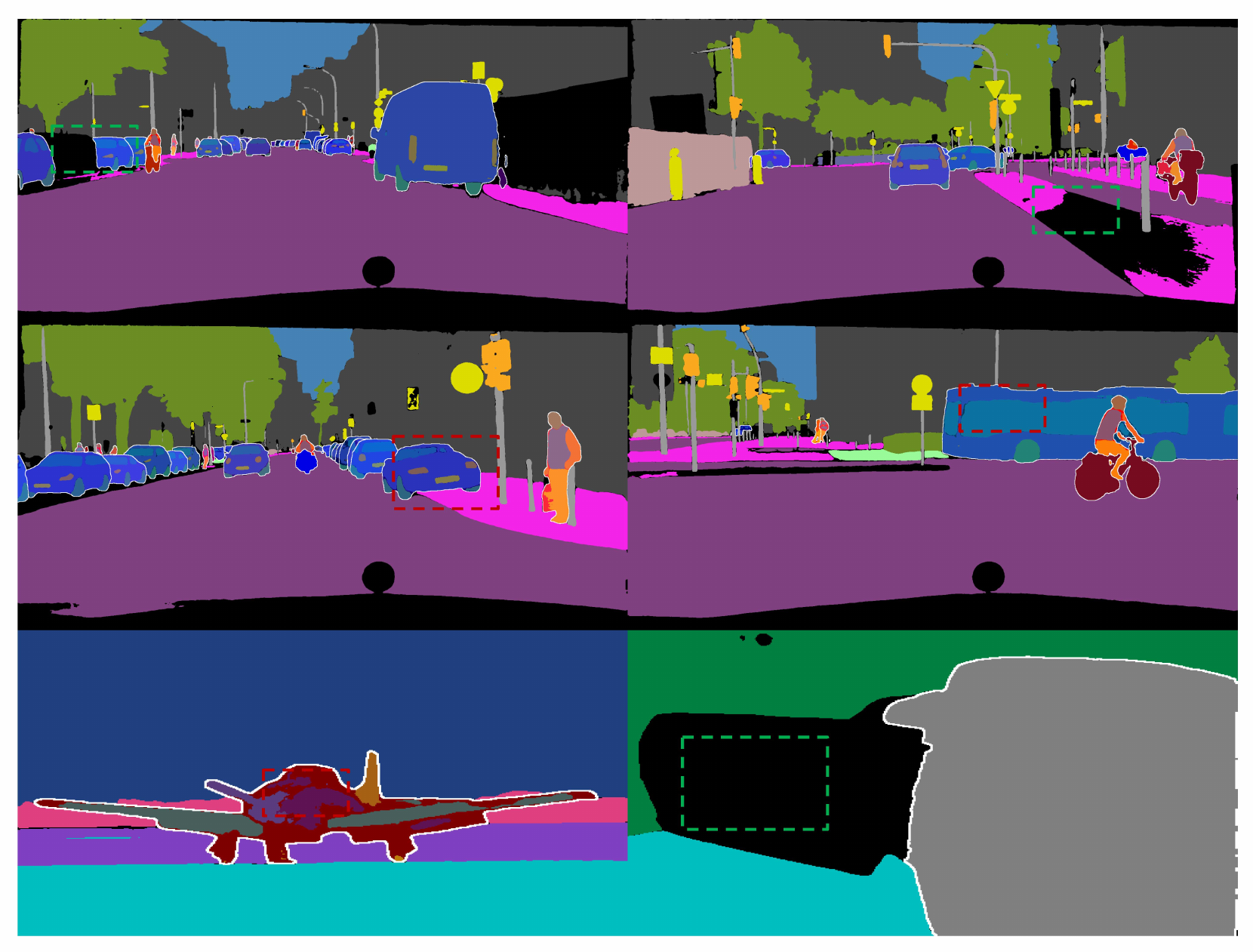}
	\caption{Failure cases on CPP (the first two rows) and PPP (the last row) datasets. Red boxes show the part segmentation inconsistency issues, and green boxes show the post-processing merging issues.}
	\label{fig:more_error}
\end{figure*}

\noindent
\textbf{More Failure Cases on Cityscapes PPS and Pascal Context PPS.} Fig.~\ref{fig:more_error} displays additional examples of errors produced by our best-performing PanopticPartFormer++ model. We hope that by presenting these examples, our findings will encourage the development of PPS methods that address two important issues: post-processing merging, where stuff is missing due to low scores (indicated in green boxes), and part segmentation inconsistency (indicated in red boxes).

	\ifCLASSOPTIONcaptionsoff
	\newpage
	\fi
	
	
	
	{
		\bibliographystyle{IEEEtran}
		\bibliography{IEEEabrv,egbib}

\begin{thebibliography}{100}
\providecommand{\url}[1]{#1}
\csname url@samestyle\endcsname
\providecommand{\newblock}{\relax}
\providecommand{\bibinfo}[2]{#2}
\providecommand{\BIBentrySTDinterwordspacing}{\spaceskip=0pt\relax}
\providecommand{\BIBentryALTinterwordstretchfactor}{4}
\providecommand{\BIBentryALTinterwordspacing}{\spaceskip=\fontdimen2\font plus
\BIBentryALTinterwordstretchfactor\fontdimen3\font minus
  \fontdimen4\font\relax}
\providecommand{\BIBforeignlanguage}[2]{{%
\expandafter\ifx\csname l@#1\endcsname\relax
\typeout{** WARNING: IEEEtran.bst: No hyphenation pattern has been}%
\typeout{** loaded for the language `#1'. Using the pattern for}%
\typeout{** the default language instead.}%
\else
\language=\csname l@#1\endcsname
\fi
#2}}
\providecommand{\BIBdecl}{\relax}
\BIBdecl

\bibitem{kirillov2019panoptic}
A.~Kirillov, K.~He, R.~Girshick, C.~Rother, and P.~Doll{\'a}r, ``Panoptic
  segmentation,'' in \emph{CVPR}, 2019.

\bibitem{liang2015human}
X.~Liang, C.~Xu, X.~Shen, J.~Yang, S.~Liu, J.~Tang, L.~Lin, and S.~Yan, ``Human
  parsing with contextualized convolutional neural network,'' in \emph{ICCV},
  2015.

\bibitem{geng2021part}
Q.~Geng, H.~Zhang, F.~Lu, X.~Huang, S.~Wang, Z.~Zhou, and R.~Yang, ``Part-level
  car parsing and reconstruction in single street view images,'' \emph{TPAMI},
  2021.

\bibitem{cordts2016cityscapes}
M.~Cordts, M.~Omran, S.~Ramos, T.~Rehfeld, M.~Enzweiler, R.~Benenson,
  U.~Franke, S.~Roth, and B.~Schiele, ``The cityscapes dataset for semantic
  urban scene understanding,'' in \emph{CVPR}, 2016.

\bibitem{degeus2021panopticparts}
D.~de~Geus, P.~Meletis, C.~Lu, X.~Wen, and G.~Dubbelman, ``Part-aware panoptic
  segmentation,'' in \emph{CVPR}, 2021.

\bibitem{maskrcnn}
K.~He, G.~Gkioxari, P.~Doll{\'a}r, and R.~Girshick, ``Mask r-cnn,'' in
  \emph{ICCV}, 2017.

\bibitem{zhao2017pyramid}
H.~Zhao, J.~Shi, X.~Qi, X.~Wang, and J.~Jia, ``Pyramid scene parsing network,''
  in \emph{CVPR}, 2017.

\bibitem{Zhao2019BSANet}
Y.~{Zhao}, J.~{Li}, Y.~{Zhang}, and Y.~{Tian}, ``{Multi-Class Part Parsing With
  Joint Boundary-Semantic Awareness},'' in \emph{ICCV}, 2019.

\bibitem{jagadeesh2022multi}
S.~K. Jagadeesh, R.~Schuster, and D.~Stricker, ``Multi-task fusion for
  efficient panoptic-part segmentation,'' in \emph{ICPRAM}, 2023.

\bibitem{xiong2019upsnet}
Y.~Xiong, R.~Liao, H.~Zhao, R.~Hu, M.~Bai, E.~Yumer, and R.~Urtasun, ``Upsnet:
  A unified panoptic segmentation network,'' in \emph{CVPR}, 2019.

\bibitem{kirillov2019panopticfpn}
A.~Kirillov, R.~Girshick, K.~He, and P.~Doll{\'a}r, ``Panoptic feature pyramid
  networks,'' in \emph{CVPR}, 2019.

\bibitem{li2020panopticFCN}
Y.~Li, H.~Zhao, X.~Qi, L.~Wang, Z.~Li, J.~Sun, and J.~Jia, ``Fully
  convolutional networks for panoptic segmentation,'' in \emph{CVPR}, 2021.

\bibitem{chen2020banet}
Y.~Chen, G.~Lin, S.~Li, O.~Bourahla, Y.~Wu, F.~Wang, J.~Feng, M.~Xu, and X.~Li,
  ``Banet: Bidirectional aggregation network with occlusion handling for
  panoptic segmentation,'' in \emph{CVPR}, 2020.

\bibitem{li2018learning}
J.~Li, A.~Raventos, A.~Bhargava, T.~Tagawa, and A.~Gaidon, ``Learning to fuse
  things and stuff,'' \emph{arXiv:1812.01192}, 2018.

\bibitem{porzi2019seamless}
L.~Porzi, S.~R. Bulo, A.~Colovic, and P.~Kontschieder, ``Seamless scene
  segmentation,'' in \emph{CVPR}, 2019.

\bibitem{yang2019sognet}
Y.~Yang, H.~Li, X.~Li, Q.~Zhao, J.~Wu, and Z.~Lin, ``Sognet: Scene overlap
  graph network for panoptic segmentation,'' in \emph{AAAI}, 2020.

\bibitem{Wu2020AutoPanopticCM}
Y.~Wu, G.~Zhang, H.~Xu, X.~Liang, and L.~Lin, ``Auto-panoptic: Cooperative
  multi-component architecture search for panoptic segmentation,'' in
  \emph{NeurIPS}, 2020.

\bibitem{cheng2020panoptic}
B.~Cheng, M.~D. Collins, Y.~Zhu, T.~Liu, T.~S. Huang, H.~Adam, and L.-C. Chen,
  ``Panoptic-deeplab: A simple, strong, and fast baseline for bottom-up
  panoptic segmentation,'' in \emph{CVPR}, 2020.

\bibitem{axialDeeplab}
H.~Wang, Y.~Zhu, B.~Green, H.~Adam, A.~Yuille, and L.-C. Chen, ``Axial-deeplab:
  Stand-alone axial-attention for panoptic segmentation,'' in \emph{ECCV},
  2020.

\bibitem{detr}
N.~Carion, F.~Massa, G.~Synnaeve, N.~Usunier, A.~Kirillov, and S.~Zagoruyko,
  ``End-to-end object detection with transformers,'' in \emph{ECCV}, 2020.

\bibitem{wang2020maxDeeplab}
H.~Wang, Y.~Zhu, H.~Adam, A.~Yuille, and L.-C. Chen, ``Max-deeplab: End-to-end
  panoptic segmentation with mask transformers,'' in \emph{CVPR}, 2021.

\bibitem{cheng2021maskformer}
B.~Cheng, A.~G. Schwing, and A.~Kirillov, ``Per-pixel classification is not all
  you need for semantic segmentation,'' in \emph{NeurIPS}, 2021.

\bibitem{zhang2021knet}
W.~Zhang, J.~Pang, K.~Chen, and C.~C. Loy, ``K-net: Towards unified image
  segmentation,'' in \emph{NeurIPS}, 2021.

\bibitem{cheng2021mask2former}
B.~Cheng, I.~Misra, A.~G. Schwing, A.~Kirillov, and R.~Girdhar,
  ``Masked-attention mask transformer for universal image segmentation,'' in
  \emph{CVPR}, 2022.

\bibitem{peize2020sparse}
P.~Sun, R.~Zhang, Y.~Jiang, T.~Kong, C.~Xu, W.~Zhan, M.~Tomizuka, L.~Li,
  Z.~Yuan, C.~Wang, and P.~Luo, ``{SparseR-CNN}: End-to-end object detection
  with learnable proposals,'' in \emph{CVPR}, 2021.

\bibitem{vaswani2017attention}
A.~Vaswani, N.~Shazeer, N.~Parmar, J.~Uszkoreit, L.~Jones, A.~N. Gomez,
  L.~Kaiser, and I.~Polosukhin, ``Attention is all you need,'' in
  \emph{NeurIPS}, 2017.

\bibitem{liu2021swin}
Z.~Liu, Y.~Lin, Y.~Cao, H.~Hu, Y.~Wei, Z.~Zhang, S.~Lin, and B.~Guo, ``Swin
  transformer: Hierarchical vision transformer using shifted windows,'' in
  \emph{ICCV}, 2021.

\bibitem{li2019attention}
Y.~Li, X.~Chen, Z.~Zhu, L.~Xie, G.~Huang, D.~Du, and X.~Wang,
  ``Attention-guided unified network for panoptic segmentation,'' in
  \emph{CVPR}, 2019.

\bibitem{li2022panopticpartformer}
X.~Li, S.~Xu, Y.~Yang, G.~Cheng, Y.~Tong, and D.~Tao, ``Panoptic-partformer:
  Learning a unified model for panoptic part segmentation,'' in \emph{ECCV},
  2022.

\bibitem{qi2018learning}
S.~Qi, W.~Wang, B.~Jia, J.~Shen, and S.-C. Zhu, ``Learning human-object
  interactions by graph parsing neural networks,'' in \emph{ECCV}, 2018.

\bibitem{liu2021unsupervised}
S.~Liu, L.~Zhang, X.~Yang, H.~Su, and J.~Zhu, ``Unsupervised part segmentation
  through disentangling appearance and shape,'' in \emph{CVPR}, 2021.

\bibitem{choudhury2021unsupervised}
S.~Choudhury, I.~Laina, C.~Rupprecht, and A.~Vedaldi, ``Unsupervised part
  discovery from contrastive reconstruction,'' in \emph{NeurIPS}, 2021.

\bibitem{fang2018weakly}
H.-S. Fang, G.~Lu, X.~Fang, J.~Xie, Y.-W. Tai, and C.~Lu, ``Weakly and semi
  supervised human body part parsing via pose-guided knowledge transfer,'' in
  \emph{CVPR}, 2018.

\bibitem{wang2019CNIF}
W.~Wang, Z.~Zhang, S.~Qi, J.~Shen, Y.~Pang, and L.~Shao, ``Learning
  compositional neural information fusion for human parsing,'' in \emph{ICCV},
  2019.

\bibitem{li2017holistic}
Q.~Li, A.~Arnab, and P.~H. Torr, ``Holistic, instance-level human parsing,''
  \emph{arXiv:1709.03612}, 2017.

\bibitem{yang2019parsing}
L.~Yang, Q.~Song, Z.~Wang, and M.~Jiang, ``Parsing {R-CNN} for instance-level
  human analysis,'' in \emph{CVPR}, 2019.

\bibitem{ji2019learning}
R.~Ji, D.~Du, L.~Zhang, L.~Wen, Y.~Wu, C.~Zhao, F.~Huang, and S.~Lyu,
  ``Learning semantic neural tree for human parsing,'' in \emph{ECCV}, 2020.

\bibitem{gong2018instance}
K.~Gong, X.~Liang, Y.~Li, Y.~Chen, M.~Yang, and L.~Lin, ``Instance-level human
  parsing via part grouping network,'' in \emph{ECCV}, 2018.

\bibitem{zhao2018understanding}
J.~Zhao, J.~Li, Y.~Cheng, T.~Sim, S.~Yan, and J.~Feng, ``Understanding humans
  in crowded scenes: Deep nested adversarial learning and a new benchmark for
  multi-human parsing,'' in \emph{ACM-MM}, 2018.

\bibitem{zhou2021differentiable}
T.~Zhou, W.~Wang, S.~Liu, Y.~Yang, and L.~Van~Gool, ``Differentiable
  multi-granularity human representation learning for instance-aware human
  semantic parsing,'' in \emph{CVPR}, 2021.

\bibitem{michieli2020gmnet}
U.~Michieli, E.~Borsato, L.~Rossi, and P.~Zanuttigh, ``{GMNet: Graph Matching
  Network for Large Scale Part Semantic Segmentation in the Wild},'' in
  \emph{ECCV}, 2020.

\bibitem{deeplabv3plus}
L.-C. Chen, Y.~Zhu, G.~Papandreou, F.~Schroff, and H.~Adam, ``Encoder-decoder
  with atrous separable convolution for semantic image segmentation,'' in
  \emph{ECCV}, 2018.

\bibitem{qiao2021detectors}
S.~Qiao, L.-C. Chen, and A.~Yuille, ``Detectors: Detecting objects with
  recursive feature pyramid and switchable atrous convolution,'' in
  \emph{CVPR}, 2021.

\bibitem{hou2020real}
R.~Hou, J.~Li, A.~Bhargava, A.~Raventos, V.~Guizilini, C.~Fang, J.~Lynch, and
  A.~Gaidon, ``Real-time panoptic segmentation from dense detections,'' in
  \emph{CVPR}, 2020.

\bibitem{li2022panopticsegformer}
Z.~Li, W.~Wang, E.~Xie, Z.~Yu, A.~Anandkumar, J.~M. Alvarez, P.~Luo, and T.~Lu,
  ``Panoptic segformer: Delving deeper into panoptic segmentation with
  transformers,'' in \emph{CVPR}, 2022.

\bibitem{yuan2022polyphonicformer}
H.~Yuan, X.~Li, Y.~Yang, G.~Cheng, J.~Zhang, Y.~Tong, L.~Zhang, and D.~Tao,
  ``Polyphonicformer: Unified query learning for depth-aware video panoptic
  segmentation,'' in \emph{ECCV}, 2022.

\bibitem{yu2022kmaxdeeplab}
Q.~Yu, H.~Wang, S.~Qiao, M.~Collins, Y.~Zhu, H.~Adam, A.~Yuille, and L.-C.
  Chen, ``k-means mask transformer,'' in \emph{ECCV}, 2022.

\bibitem{felzenszwalb2010object}
P.~F. Felzenszwalb, R.~B. Girshick, D.~McAllester, and D.~Ramanan, ``Object
  detection with discriminatively trained part-based models,'' \emph{TPAMI},
  2010.

\bibitem{fergus2003object}
R.~Fergus, P.~Perona, and A.~Zisserman, ``Object class recognition by
  unsupervised scale-invariant learning,'' in \emph{CVPR}, 2003.

\bibitem{zhang2013deformable}
N.~Zhang, R.~Farrell, F.~Iandola, and T.~Darrell, ``Deformable part descriptors
  for fine-grained recognition and attribute prediction,'' in \emph{CVPR},
  2013.

\bibitem{tritrong2021repurposing}
N.~Tritrong, P.~Rewatbowornwong, and S.~Suwajanakorn, ``Repurposing gans for
  one-shot semantic part segmentation,'' in \emph{CVPR}, 2021.

\bibitem{zhang2021datasetgan}
Y.~Zhang, H.~Ling, J.~Gao, K.~Yin, J.-F. Lafleche, A.~Barriuso, A.~Torralba,
  and S.~Fidler, ``Datasetgan: Efficient labeled data factory with minimal
  human effort,'' in \emph{CVPR}, 2021.

\bibitem{sabour2021unsupervised}
S.~Sabour, A.~Tagliasacchi, S.~Yazdani, G.~Hinton, and D.~J. Fleet,
  ``Unsupervised part representation by flow capsules,'' in \emph{ICML}, 2021.

\bibitem{xu2019unsupervised}
Z.~Xu, Z.~Liu, C.~Sun, K.~Murphy, W.~T. Freeman, J.~B. Tenenbaum, and J.~Wu,
  ``Unsupervised discovery of parts, structure, and dynamics,''
  \emph{arXiv:1903.05136}, 2019.

\bibitem{yang2020eccv}
L.~Yang, Q.~Song, Z.~Wang, M.~Hu, C.~Liu, X.~Xin, W.~Jia, and S.~Xu,
  ``Renovating parsing {R-CNN} for accurate multiple human parsing,'' in
  \emph{ECCV}, 2020.

\bibitem{Everingham2010Pascal}
M.~Everingham, L.~Van~Gool, C.~K. Williams, J.~Winn, and A.~Zisserman, ``{The
  Pascal Visual Object Classes (VOC) Challenge},'' \emph{IJCV}, 2010.

\bibitem{tang2022visual}
C.~Tang, L.~Xie, X.~Zhang, X.~Hu, and Q.~Tian, ``Visual recognition by
  request,'' \emph{arXiv:2207.14227}, 2022.

\bibitem{VIT}
A.~Dosovitskiy, L.~Beyer, A.~Kolesnikov, D.~Weissenborn, X.~Zhai,
  T.~Unterthiner, M.~Dehghani, M.~Minderer, G.~Heigold, S.~Gelly \emph{et~al.},
  ``An image is worth 16x16 words: Transformers for image recognition at
  scale,'' \emph{arXiv:2010.11929}, 2020.

\bibitem{deit_vit}
H.~Touvron, M.~Cord, M.~Douze, F.~Massa, A.~Sablayrolles, and H.~J{\'e}gou,
  ``Training data-efficient image transformers \& distillation through
  attention,'' in \emph{ICML}, 2021.

\bibitem{zhang2022eatformer}
J.~Zhang, X.~Li, Y.~Wang, C.~Wang, Y.~Yang, Y.~Liu, and D.~Tao, ``Eatformer:
  improving vision transformer inspired by evolutionary algorithm,''
  \emph{arXiv:2206.09325}, 2022.

\bibitem{li2022uniformer}
K.~Li, Y.~Wang, P.~Gao, G.~Song, Y.~Liu, H.~Li, and Y.~Qiao, ``Uniformer:
  Unified transformer for efficient spatiotemporal representation learning,''
  \emph{TPAMI}, 2022.

\bibitem{guo2021cmt}
J.~Guo, K.~Han, H.~Wu, C.~Xu, Y.~Tang, C.~Xu, and Y.~Wang, ``Cmt: Convolutional
  neural networks meet vision transformers,'' in \emph{CVPR}, 2022.

\bibitem{MaskedAutoencoders2021}
K.~He, X.~Chen, S.~Xie, Y.~Li, P.~Doll{\'a}r, and R.~Girshick, ``Masked
  autoencoders are scalable vision learners,'' \emph{arXiv:2111.06377}, 2021.

\bibitem{CLIP}
A.~Radford, J.~W. Kim, C.~Hallacy, A.~Ramesh, G.~Goh, S.~Agarwal, G.~Sastry,
  A.~Askell, P.~Mishkin, J.~Clark \emph{et~al.}, ``Learning transferable visual
  models from natural language supervision,'' in \emph{ICML}, 2021.

\bibitem{OpenSeg}
G.~Ghiasi, X.~Gu, Y.~Cui, and T.-Y. Lin, ``Scaling open-vocabulary image
  segmentation with image-level labels,'' in \emph{ECCV}, 2022.

\bibitem{chen2022vitadapter}
Z.~Chen, Y.~Duan, W.~Wang, J.~He, T.~Lu, J.~Dai, and Y.~Qiao, ``Vision
  transformer adapter for dense predictions,'' in \emph{ICLR}, 2023.

\bibitem{zhu2020deformabledetr}
X.~Zhu, W.~Su, L.~Lu, B.~Li, X.~Wang, and J.~Dai, ``Deformable detr: Deformable
  transformers for end-to-end object detection,'' in \emph{ICLR}, 2020.

\bibitem{dong2021solq}
B.~Dong, F.~Zeng, T.~Wang, X.~Zhang, and Y.~Wei, ``Solq: Segmenting objects by
  learning queries,'' in \emph{NeurIPS}, 2021.

\bibitem{xu2022fashionformer}
S.~Xu, X.~Li, J.~Wang, G.~Cheng, Y.~Tong, and D.~Tao, ``Fashionformer: A
  simple, effective and unified baseline for human fashion segmentation and
  recognition,'' in \emph{ECCV}, 2022.

\bibitem{zhou2022transvod}
Q.~Zhou, X.~Li, L.~He, Y.~Yang, G.~Cheng, Y.~Tong, L.~Ma, and D.~Tao,
  ``Transvod: End-to-end video object detection with spatial-temporal
  transformers,'' \emph{TPAMI}, 2022.

\bibitem{meinhardt2021trackformer}
T.~Meinhardt, A.~Kirillov, L.~Leal-Taixe, and C.~Feichtenhofer, ``Trackformer:
  Multi-object tracking with transformers,'' in \emph{CVPR}, 2022.

\bibitem{li2022videoknet}
X.~Li, W.~Zhang, J.~Pang, K.~Chen, G.~Cheng, Y.~Tong, and C.~C. Loy, ``Video
  k-net: A simple, strong, and unified baseline for video segmentation,'' in
  \emph{CVPR}, 2022.

\bibitem{densePredic_dac_2018}
K.~Tateno, N.~Navab, and F.~Tombari, ``Distortion-aware convolutional filters
  for dense prediction in panoramic images,'' \emph{ECCV}, 2018.

\bibitem{Pad-net_2018}
D.~Xu, W.~Ouyang, X.~Wang, and N.~Sebe, ``Pad-net: Multi-tasks guided
  prediction-and-distillation network for simultaneous depth estimation and
  scene parsing,'' in \emph{CVPR}, 2018.

\bibitem{PAP-Net_2019}
Z.~Zhang, Z.~Cui, C.~Xu, Y.~Yan, N.~Sebe, and J.~Yang, ``Pattern-affinitive
  propagation across depth, surface normal and semantic segmentation,'' in
  \emph{CVPR}, 2019.

\bibitem{densePredic_cwkdis_2021}
C.~Shu, Y.~Liu, J.~Gao, Z.~Yan, and C.~Shen, ``Channel-wise knowledge
  distillation for dense prediction,'' \emph{ICCV}, 2021.

\bibitem{densePredic_dcmd_2021}
N.~Takahashi and Y.~Mitsufuji, ``Densely connected multi-dilated convolutional
  networks for dense prediction tasks,'' in \emph{CVPR}, 2021.

\bibitem{pattern_struct_diffusion_2020}
Z.~Ling, C.~Zhen, X.~Chunyan, Z.~Zhenyu, W.~Chaoqun, Z.~Tong, and Y.~Jian,
  ``Pattern-structure diffusion for multi-task learning,'' \emph{CVPR}, 2020.

\bibitem{multitask_mtst_2021}
G.~Ghiasi, B.~Zoph, E.~D. Cubuk, Q.~V. Le, and T.-Y. Lin, ``Multi-task
  self-training for learning general representations,'' in \emph{ICCV}, 2021.

\bibitem{multitask_UM_2019}
J.~N. Kundu, N.~Lakkakula, and R.~V. Babu, ``Um-adapt: Unsupervised multi-task
  adaptation using adversarial cross-task distillation,'' in \emph{ICCV}, 2019.

\bibitem{Mti-net_2020}
S.~Vandenhende, S.~Georgoulis, and L.~Van~Gool, ``Mti-net: Multi-scale task
  interaction networks for multi-task learning,'' in \emph{ECCV}, 2020.

\bibitem{invpt2022}
H.~Ye and D.~Xu, ``Inverted pyramid multi-task transformer for dense scene
  understanding,'' in \emph{ECCV}, 2022.

\bibitem{xu2023mqformer}
Y.~Xu, X.~Li, H.~Yuan, Y.~Yang, and L.~Zhang, ``Multi-task learning with
  multi-query transformer for dense prediction,'' \emph{IEEE-TCSVT}, 2023.

\bibitem{taskprompter2023}
H.~Ye and D.~Xu, ``Taskprompter: Spatial-channel multi-task prompting for dense
  scene understanding,'' in \emph{ICLR}, 2023.

\bibitem{resnet}
K.~{He}, X.~{Zhang}, S.~{Ren}, and J.~{Sun}, ``Deep residual learning for image
  recognition,'' in \emph{CVPR}, 2016.

\bibitem{fpn}
T.-Y. {Lin}, P.~{Dollár}, R.~B. {Girshick}, K.~{He}, B.~{Hariharan}, and S.~J.
  {Belongie}, ``Feature pyramid networks for object detection,'' in
  \emph{CVPR}, 2017.

\bibitem{sfnet}
X.~Li, A.~You, Z.~Zhu, H.~Zhao, M.~Yang, K.~Yang, and Y.~Tong, ``Semantic flow
  for fast and accurate scene parsing,'' in \emph{ECCV}, 2020.

\bibitem{wang2020solo}
X.~Wang, T.~Kong, C.~Shen, Y.~Jiang, and L.~Li, ``{SOLO}: Segmenting objects by
  locations,'' in \emph{ECCV}, 2020.

\bibitem{wang2020solov2}
X.~Wang, R.~Zhang, T.~Kong, L.~Li, and C.~Shen, ``{SOLOv2}: Dynamic and fast
  instance segmentation,'' in \emph{NeurIPS}, 2020.

\bibitem{coco_dataset}
T.-Y. Lin, M.~Maire, S.~Belongie, J.~Hays, P.~Perona, D.~Ramanan,
  P.~Doll{\'a}r, and C.~L. Zitnick, ``Microsoft coco: Common objects in
  context,'' in \emph{ECCV}, 2014.

\bibitem{dice_loss}
F.~Milletari, N.~Navab, and S.~Ahmadi, ``{V-Net: Fully} convolutional neural
  networks for volumetric medical image segmentation,'' in \emph{3DV}, 2016.

\bibitem{bao2021beit}
H.~Bao, L.~Dong, and F.~Wei, ``Beit: Bert pre-training of image transformers,''
  \emph{arXiv:2106.08254}, 2021.

\bibitem{xavier_init}
X.~Glorot and Y.~Bengio, ``Understanding the difficulty of training deep
  feedforward neural networks,'' in \emph{AISTATS}, 2010.

\bibitem{ADAMW}
I.~Loshchilov and F.~Hutter, ``Decoupled weight decay regularization,'' in
  \emph{ICLR}, 2019.

\bibitem{xie2021segformer}
E.~Xie, W.~Wang, Z.~Yu, A.~Anandkumar, J.~M. Alvarez, and P.~Luo, ``Segformer:
  Simple and efficient design for semantic segmentation with transformers,'' in
  \emph{NeurIPS}, 2021.

\bibitem{mohan2021efficientps}
R.~Mohan and A.~Valada, ``Efficientps: Efficient panoptic segmentation,''
  \emph{IJCV}, 2021.

\bibitem{tan2019efficientnet}
M.~Tan and Q.~Le, ``Efficientnet: Rethinking model scaling for convolutional
  neural networks,'' in \emph{ICML}, 2019.

\bibitem{ocrnet}
Y.~Yuan, X.~Chen, and J.~Wang, ``Object-contextual representations for semantic
  segmentation,'' in \emph{ECCV}, 2020.

\bibitem{wang2020deep}
J.~Wang, K.~Sun, T.~Cheng, B.~Jiang, C.~Deng, Y.~Zhao, D.~Liu, Y.~Mu, M.~Tan,
  X.~Wang \emph{et~al.}, ``Deep high-resolution representation learning for
  visual recognition,'' \emph{TPAMI}, 2020.

\bibitem{liang2020polytransform}
J.~Liang, N.~Homayounfar, W.-C. Ma, Y.~Xiong, R.~Hu, and R.~Urtasun,
  ``Polytransform: Deep polygon transformer for instance segmentation,'' in
  \emph{CVPR}, 2020.

\bibitem{zhang2020resnest}
H.~Zhang, C.~Wu, Z.~Zhang, Y.~Zhu, H.~Lin, Z.~Zhang, Y.~Sun, T.~He, J.~Mueller,
  R.~Manmatha \emph{et~al.}, ``Resnest: Split-attention networks,'' in
  \emph{CVPR}, 2022.

\bibitem{chollet2017xception}
F.~Chollet, ``Xception: Deep learning with depthwise separable convolutions,''
  in \emph{CVPR}, 2017.

\bibitem{neuhold2017mapillary}
G.~Neuhold, T.~Ollmann, S.~Rota~Bulo, and P.~Kontschieder, ``The mapillary
  vistas dataset for semantic understanding of street scenes,'' in \emph{ICCV},
  2017.

\bibitem{liu2022convnet}
Z.~Liu, H.~Mao, C.-Y. Wu, C.~Feichtenhofer, T.~Darrell, and S.~Xie, ``A convnet
  for the 2020s,'' in \emph{CVPR}, 2022.

\bibitem{deeplabv3}
L.-C. Chen, G.~Papandreou, F.~Schroff, and H.~Adam, ``Rethinking atrous
  convolution for semantic image segmentation,'' \emph{arXiv:1706.05587}, 2017.

\bibitem{deformablev2}
X.~Zhu, H.~Hu, S.~Lin, and J.~Dai, ``Deformable convnets v2: More deformable,
  better results,'' in \emph{CVPR}, 2019.

\bibitem{li2020unifying}
Q.~Li, X.~Qi, and P.~H. Torr, ``Unifying training and inference for panoptic
  segmentation,'' in \emph{CVPR}, 2020.

\bibitem{li2021fully}
Y.~Li, H.~Zhao, X.~Qi, Y.~Chen, L.~Qi, L.~Wang, Z.~Li, J.~Sun, and J.~Jia,
  ``Fully convolutional networks for panoptic segmentation with point-based
  supervision,'' \emph{TPAMI}, 2022.

\bibitem{milletari2016v}
F.~Milletari, N.~Navab, and S.-A. Ahmadi, ``V-net: Fully convolutional neural
  networks for volumetric medical image segmentation,'' in \emph{3DV}, 2016.

\bibitem{li2020improving}
X.~Li, X.~Li, L.~Zhang, G.~Cheng, J.~Shi, Z.~Lin, S.~Tan, and Y.~Tong,
  ``Improving semantic segmentation via decoupled body and edge supervision,''
  in \emph{ECCV}, 2020.

\bibitem{kirillov2019pointrend}
A.~Kirillov, Y.~Wu, K.~He, and R.~Girshick, ``Pointrend: image segmentation as
  rendering. arxiv,'' in \emph{CVPR}, 2020.

\bibitem{yu2020bdd100k}
F.~Yu, H.~Chen, X.~Wang, W.~Xian, Y.~Chen, F.~Liu, V.~Madhavan, and T.~Darrell,
  ``Bdd100k: A diverse driving dataset for heterogeneous multitask learning,''
  in \emph{CVPR}, 2020.

\end{thebibliography}
	}

\end{document}